\documentclass[10pt,journal,compsoc]{IEEEtran}

\ifCLASSOPTIONcompsoc
    \usepackage[noadjust]{cite}
\else
  \usepackage[noadjust]{cite}
\fi
\ifCLASSINFOpdf  
\else  
\fi
\usepackage{amsmath}
\usepackage{url}
\usepackage{xcolor}
\usepackage{times}
\usepackage{epsfig}
\usepackage{graphicx}
\usepackage{caption}
\usepackage{float}
\usepackage{amssymb} 
\usepackage{footmisc}
\usepackage{lineno}
\usepackage{color}
\usepackage{subfigure}
\usepackage{multirow}
\usepackage{dsfont}
\usepackage{mathtools}
\usepackage{setspace}
\usepackage{csquotes}
\usepackage{breqn}
\usepackage{lettrine}
\usepackage{upquote}
\usepackage{adjustbox}
\usepackage{wrapfig}
\usepackage{lipsum}
\usepackage{mathtools}
\usepackage{comment}
\usepackage{bm}
\usepackage{array}
\usepackage{tabu}
\usepackage{booktabs}
\usepackage{enumerate}
\DeclareCaptionFormat{myformat}{\fontsize{8}{10}\selectfont#1#2#3}
\DeclareMathAlphabet\mathbfcal{OMS}{cmsy}{b}{n}

\DeclareMathAlphabet{\pazocal}{OMS}{zplm}{m}{n}
\DeclareMathAlphabet{\mathpzc}{OT1}{pzc}{m}{it}

\usepackage[ruled,vlined]{algorithm2e}
\usepackage[colorlinks=true,linkcolor=blue]{hyperref}
\def\arrvline{\hfil\kern\arraycolsep\vline\kern-\arraycolsep\hfilneg}

\ifCLASSINFOpdf
\else
\fi

\begin{document}
\bstctlcite{IEEEexample:BSTcontrol}
\title{Improving Underwater Visual Tracking With a Large Scale Dataset and Image Enhancement}

\author{Basit Alawode,
Fayaz Ali Dharejo,
Mehnaz Ummar,
Yuhang Guo,
Arif Mahmood,
Naoufel Werghi,~\IEEEmembership{Senior Member,~IEEE,}
Fahad Shahbaz Khan, ~\IEEEmembership{IEEE Senior Member},
~Jiri Matas, ~\IEEEmembership{IEEE Senior Member}, and
Sajid Javed
\IEEEcompsocitemizethanks{\IEEEcompsocthanksitem  B. \ Alawode, F.\ A.\ Dharejo, M. Ummar, Y. Guo, N.\ Werghi, and S.\ Javed are with the EECS Department, Khalifa University of Science and Technology, P.O Box: 127788, Abu Dhabi, UAE. (email: sajid.javed@ku.ac.ae).
\IEEEcompsocthanksitem F.\ S.\ Khan is with the computer vision department, MBZUAI, Abu Dhabi, UAE.
\IEEEcompsocthanksitem J.\ Matas is with Center for Machine Perception, Czech Technical University, Prague.}}

\markboth{Journal of \LaTeX\ Class Files,~Vol.~14, No.~8, August~2015}
{Javed \etal: VOT}
\IEEEtitleabstractindextext{%


\begin{abstract}
%
This paper presents a new dataset and general tracker enhancement method for Underwater Visual Object Tracking (UVOT).
Despite its significance, underwater tracking has remained unexplored due to data inaccessibility. It poses distinct challenges; the underwater environment exhibits non-uniform lighting conditions, low visibility, lack of sharpness, low contrast, camouflage, and reflections from suspended particles.
Performance of traditional tracking methods designed primarily for terrestrial or open-air scenarios drops in such conditions.
We address the problem by proposing a novel underwater image enhancement algorithm  designed specifically to boost tracking quality. The method has resulted in a significant performance improvement, of up to 5.0$\%$ AUC, of state-of-the-art (SOTA) visual trackers.
To develop robust and accurate UVOT methods, large-scale datasets are required.
To this end, we introduce a large-scale UVOT benchmark dataset consisting of 400 video segments and 275,000 manually annotated frames enabling underwater training and evaluation of deep trackers. 
The videos are labelled with several underwater-specific tracking attributes including watercolor variation, target distractors, camouflage, target relative size, and low visibility conditions.
%
%
The UVOT400 dataset, tracking results, and the code are publicly available on: \href{https://github.com/BasitAlawode/UWVOT400}{https://github.com/BasitAlawode/UWVOT400}.
 
\end{abstract}

\begin{IEEEkeywords}
Visual Object Tracking, Underwater Tracking, Dataset, Underwater Image Enhancement, Vision Transformer.
\end{IEEEkeywords}}

\maketitle

\IEEEdisplaynontitleabstractindextext

\IEEEpeerreviewmaketitle

\IEEEraisesectionheading{\section{Introduction}
\label{sec:introduction}}
\IEEEPARstart{V}{isual} Object Tracking (VOT) is one of the fundamental and long-standing problems in computer vision \cite{javed2022visual}.
The main aim of VOT is to estimate the location of the generic moving target object in a video sequence, given its position in the first frame \cite{javed2022visual}.
It is quite challenging to learn a class-agnostic, robust-to-noise, and distractor-aware target appearance model in the presence of occlusion, lighting variations, target rotation, scale variations, and fast motion etc. \cite{javed2022visual, fan2021lasot,muller2018trackingnet }.

VOT has numerous applications such as video surveillance, autonomous driving, robotics manipulation, and navigation, sports video analysis, and human activity recognition \cite{shih2017survey, xie2021deep,kong2022human, seok2020rovins, sun2020scalability}.
In the past decade, the tracking community has considerably progressed and many end-to-end deep learning based trackers have been proposed \cite{danelljan2019atom, bhat2019learning, danelljan2020probabilistic,yan2021alpha, chen2022siamban, chen2020siamese, bhat2020learning, bertinetto2016fully}.
One of the main reasons behind this progress is the availability of a variety of large-scale open-air tracking datasets such as GOT-10K \cite{huang2019got}, LaSOT \cite{fan2021lasot}, and TrackingNet \cite{muller2018trackingnet} which are used to train and evaluate new trackers.

Underwater Visual Object Tracking (UVOT) has largely remained an unexplored research area in the computer vision community despite its numerous applications such as search and rescue operations \cite{manzanilla2019autonomous}, homeland and maritime security \cite{gonzalez2023survey}, ocean exploration \cite{boulais2020fathomnet}, sea life monitoring \cite{akkaynak2019sea}, fisheries management and control \cite{betancourt2020integrated, saleh2022computer}, underwater waste cleaning \cite{cheng2021flow}, and underwater robotics intervention \cite{8706541}.

\begin{figure}[t!]
\centering
\includegraphics[width=\linewidth]{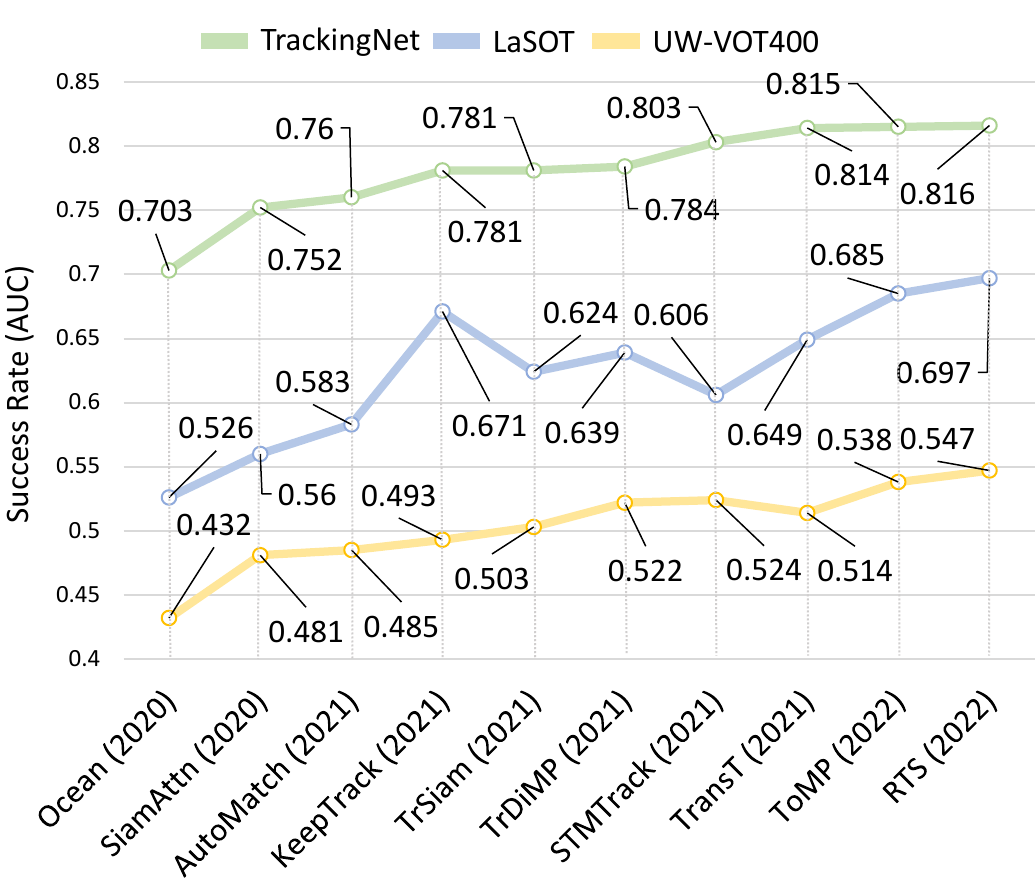}
\caption{ Underwater performance drops for all selected recent SOTA trackers compared to open-air scenarios \cite{muller2018trackingnet, fan2021lasot}, in terms of the success rate.
Trackers tested: Ocean \cite{zhang2020ocean}, SiamAttn \cite{yu2020deformable}, AutoMatch \cite{zhang2021learn}, KeepTrack \cite{mayer2021learning}, TrSiam \cite{wang2021transformer}, TrDiMP \cite{wang2021transformer}, STMTrack \cite{fu2021stmtrack}, TransT \cite{chen2021transformer}, ToMP \cite{mayer2022transforming}, and RTS~\cite{paul2022robust}.
}
\label{fig1}
\end{figure}

Performance of all tested SOTA visual trackers significantly degrades when employed in UnderWater (UW) scenes compared to their performance on open-air datasets (Fig. \ref{fig1}).
This is likely since underwater environment poses a very unique set of challenges that make UVOT very challenging resulting in deteriorated tracking performance. 
These challenges include low light and poor visibility conditions, light scattering by particles, absorption of light, non-uniform illumination conditions, target blurring, watercolor variations (such as blueish or greenish, etc.), flickering of caustic patterns \cite{liu2020real, gonzalez2023survey}, a large number of similar distractors, quick target visibility decay with increasing distance from the camera.
In an experiment performed on 25 existing SOTA trackers, we observed significant performance degradation when open-air trained trackers are employed for underwater scenarios (Fig. \ref{fig1}).
Almost all trackers have shown a significant performance gap between open-air and underwater tracking.

We empirically show that image enhancement methods improve the tracking performance by reducing the adverse effect of the UVOT specific challenges. 
For this purpose, we employ existing SOTA underwater image enhancement methods and demonstrate improvement in the UVOT performance. 
All of the SOTA trackers included in this paper have consistently obtained better tracking performance when applied to the enhanced images.
In this work, we propose a novel underwater image enhancement algorithm for the purpose of the UVOT performance improvement.

The literature dominantly deals with  open-air VOT. 
UVOT has sparsely been investigated, which is likely to be attributed due to the lack of large-scale UVOT datasets, benchmarking, and the difficulty of obtaining UW imagery.
We discuss the challenges posed by the UW environment for VOT and present a new UVOT dataset and then benchmark 25 existing SOTA trackers.

\vspace{1ex}
\noindent  \textbf{The UVOT400 Dataset:} For real-world applications of UVOT, a large, long-term tracking benchmark is required. 
We thus introduce an UW dataset, called UVOT400, containing 275,000 annotated frames with accurate bounding boxes and  an average sequence length of 688 frames.
The dataset mostly contains natural UW sequences under varying visibility conditions, captured using both moving and stationary cameras.
The dataset contains 17 different tracking attributes including UW-specific ones such as UW visibility, watercolor variations, swarm distractors, camouflage, and partial target information.
The dataset contains several diverse object categories and it thus presents the tracking community diverse underwater targets.


\vspace{1ex}
\noindent \textbf{UVOT400 Benchmark:} 
To facilitate development of underwater trackers, we establish a novel benchmark on the UVOT400 dataset.
We adopt performance measures from LaSOT \cite{fan2021lasot} and establish an extensive baseline by evaluating the performance of 25 recent open-air-trained deep learning trackers.
All show a performance drop, see Fig. \ref{fig1}.
For example, sequences with UW visibility challenges exhibit maximum degradation -  a 20$\%$ performance drop.  Similarly, sequences with UW distractors, camouflage, and watercolor challenges show 15$\%$, 12$\%$, and 10$\%$ tracking performance degradation w.r.t. open-air sequences .

\vspace{1ex}
\noindent \textbf{Underwater Image Enhancement for Tracking:} We are first to show that even standard image enhancements methods, developed to facilitate human-based image interpretation of underwater images, improve underwater tracking performance of open-air trackers. Moreover, we develop a novel, transformer based image enhancement method 
specifically designed to improve underwater tracking.
The proposed novel algorithm, called UWIE-TR (Underwater Image Enhancement for TRacking), when integrated as a pre-processing step of the SOTA trackers,
significantly boosts performance in terms of both reference-based and no-reference evaluation metrics. 

The UWIE-TR is trained on pseudo-ground truth generated by the best result, selected by expert human annotators, from a pool of outputs of commonly used image enhancement methods.
A transformer encoder is used to project the input UW images to a latent space where we perform information fusion across multiple enhanced versions of the UW images.
An image decoder is then employed to generate the enhanced images. 
The architecture is trained in an end-to-end manner minimizing loss between pseudo-ground truth and the generated images.
The evaluation of the UWIE-TR algorithm is performed using four publicly available image enhancement datasets.

\vspace{1ex}
\noindent \textbf{The Main Contributions} of this work are summarized as follows:

\begin{itemize}
\item A large and diverse high-quality UVOT400 benchmark dataset is presented, consisting of 400 sequences and 275,000 manually annotated bounding-box frames, introducing 17 distinct tracking attributes with diverse underwater creatures as targets.
\item A large-scale benchmarking of 25 recent SOTA trackers is performed on the proposed dataset, adopting established performance metrics (\cite{fan2021lasot}).
\item An UWIE-TR algorithm is introduced. It improves the UVOT performance of SOTA open-air trackers on underwater sequences. 
\item The selected SOTA trackers are re-trained on the enhanced version of the proposed dataset resulting in significant performance improvement across all compared trackers.
\end{itemize}

This work extends an earlier conference version \cite{alawode2022utb180}. 
The proposed UWIE-TR algorithm that improves the performance of the SOTA trackers is a novelty w.r.t to the conference paper.  The other new  contributions are: the number of videos is extended from 180 to 400, the number of densely annotated frames is increased from 58,000 to 275,000, the number of object categories is increased from 30 to 50, and the number of tracking attributes is also increased from 10 to 17. 
 The evaluation of SOTA trackers is extended, from from 15 to 25 trackers. 
 

The rest of this work is organized as follows: Section \ref{sec:relatedwork} provides a literature review on the open-air datasets, UVOT datasets, and recent progress on VOT.
Section \ref{sec::dataset} discusses our proposed UVOT dataset in detail. 
Section \ref{sec:UIE} presents the proposed UWIE-TR algorithm followed by a comprehensive evaluation of UW enhancement methods.
Rigorous benchmarking and experimental evaluations are presented in Sec. \ref{sec:results} while the conclusion and future research directions are shown in Sec. \ref{sec:conclusion}.

\section{Literature Review}
\label{sec:relatedwork}
The main emphasis of this work is on UVOT and benchmarking.
For completeness purposes, we also provide an overview of VOT open-air datasets and recent progress in VOT.
The literature review is categorized into the following three sections.

\subsection{Open-air VOT Datasets}
In the past decade, many open-air VOT benchmark datasets have been proposed in the literature \cite{fan2021lasot, muller2018trackingnet, huang2019got}.
These datasets can be divided into short-term and long-term datasets.
According to Fan \textit{et al.}, VOT sequences having an average length larger than 600 frames may be considered long-term datasets \cite{fan2021lasot}.
The popular short-term datasets include OTB13 \cite{wu2013online}, OTB15 \cite{7001050}, TC128 \cite{7277070}, VOT 2014-2022 series \cite{VOT2014, VOT2015, VOT2016, VOT2017, VOT2018, VOT2019, VOT2020, kristan2022tenth}, GOT-10k \cite{huang2019got}, NUS-PRO \cite{li2015nus}, and TrackingNet \cite{muller2018trackingnet}.
The popular long-term datasets include UAV123 \cite{mueller2016benchmark}, UAV20L \cite{mueller2016benchmark}, LaSOT \cite{fan2019lasot, fan2021lasot}, NFS \cite{kiani2017need}, CDTB \cite{lukezic2019cdtb}, and VOT-LT 2018-2022 series \cite{VOT2018, VOT2019, VOT2020,VOT2021, VOT2022}.
These datasets are also summarized in Table \ref{table1}.

The following are the main reasons behind the advancement of SOTA tracking performance.
There are several \textit{online evaluation platforms} provided by the original authors to facilitate standardized comparisons among different trackers.
For instance, TrackingNet, Got-10K, and VOT have provided their own evaluation platforms.
Each dataset provides distinct \textit{tracking attributes} such as occlusion, motion blur, lighting condition, scale variations, etc. to name a few \cite{huang2019got, fan2021lasot}.
The attribute labeling has been provided at the video level to facilitate attribute-level comparisons among different trackers.
Several datasets publicly maintain their \textit{performance leaderboards} such as TrackingNet \footnote{https://eval.ai/web/challenges/challenge-page/1805/evaluation}, GOT-10K  \footnote{http://got-10k.aitestunion.com/leaderboard}, and VOT \cite{VOT2014, VOT2015, VOT2016, VOT2017, VOT2018, VOT2019, VOT2020, VOT2021, VOT2022}.
The presence of these leaderboards further facilitates performance comparisons and updates the tracking community against the best-performing results on a particular dataset.
\textit{Annual VOT contests and challenges} also motivate the community to come up with better tracking solutions.
For example, the VOT community have been organizing these contests since 2014 \cite{VOT2014, VOT2022}.
Many authors have suggested using external object detection datasets for learning to improve \textit{robust feature representation} for better tracking \cite{javed2022visual}.
However, large-scale tracking datasets such as LaSOT etc., have significantly improved robust class-agnostic feature representation learning.
Unfortunately, no such efforts have been contributed towards underwater VOT.

\begin{table*}[t!]
\caption{Comparison of the proposed UVOT400 dataset and the most popular Open-Air (OA) benchmarks. ``Eva.'' and ``Tra.'' stands for  evaluation and training.}
\begin{center}
\makebox[\linewidth]{
\scalebox{0.74}{
\begin{tabu}{|c|c|c|c|c|c|c|c|c|c|c|c|c|c|}
\tabucline[1.5pt]{-}
Statistics&OTB-13&OTB-15&TC-128&CDTB&NUS-PRO&UAV123&UAV20L&VOT2022&GOT-10K&LaSOT&TrackingNet&UOT100&UVOT \\
&\cite{wu2013online}&\cite{7001050}&\cite{7277070}&\cite{lukezic2019cdtb}&\cite{li2015nus}&\cite{mueller2016benchmark}&\cite{mueller2016benchmark}&\cite{kristan2022tenth}&\cite{huang2019got}&\cite{fan2021lasot}&\cite{muller2018trackingnet} &\cite{panetta2021comprehensive}&Proposed\\\tabucline[1.5pt]{-}
Type&OA&OA&OA&OA&OA&OA&OA&OA&OA&OA&OA&UW&UW\\\tabucline[0.5pt]{-}
No. of videos&51&100&128&80&365&123&20&60&10K&1550&30.643K&104&400\\\tabucline[0.5pt]{-}
Minimum frames&71&71&71&406&146&109&1717&41&29&1K&96&264&40\\\tabucline[0.5pt]{-}
Maximum frames&3872&3872&3872&2501&5040&3085&5527&1500&1418&11397&2368&1764&3273\\\tabucline[0.5pt]{-}
Average frames&578&590&429&1274&371&915&3830&356&149&2502&471&698&688\\\tabucline[0.5pt]{-}
Boxes&29K&59K&55K&102K&135K&113K&59K&21K&1.45M&3.87M&14M&74K&275K\\\tabucline[0.5pt]{-}
Object classes&10&16&27&23&8&9&5&24&563&85&21&Unknown&50\\\tabucline[0.5pt]{-}
No. of Attributes&11&11&11&13&n/a&12&12&n/a&6&14&15&3&17\\\tabucline[0.5pt]{-}
Framerate&30fps&30fps&30fps&30fps&30fps&30fps&30fps&30fps&10fps&130fps&130fps&30fps&30fps\\\tabucline[0.5pt]{-}
Objective&Eva.&Eva.&Eva.&Eva.&Eva.&Eva.&Eva.&Eva.&Eva./Tra.&Eva./Tra.&Eva./Tra.&Eva.&Eva./Tra.\\\tabucline[1.5pt]{-}

\end{tabu}
}
}
\end{center}
\label{table1}
\end{table*}

\subsection{Underwater VOT Datasets}
Compared to open-air tracking benchmarks, underwater VOT datasets are scarcely available. 
To the best of our knowledge, UOT32 \cite{kezebou2019underwater}, UOT100 \cite{panetta2021comprehensive}, UTB180 \cite{alawode2022utb180}, and VMAT \cite{cai2023semi} are the only available UW tracking datasets where UOT100 is the extended version of UOT32.
The proposed UVOT dataset in this work is an extended version of UTB180.
The UOT100 dataset consists of 104 sequences with 74,000 annotated bounding boxes obtained from YouTube. 
It captures many underwater distortions and non-uniform lighting conditions. 
A new dataset has recently been proposed as UTB180 consisting of 180 sequences with 58,000 annotated frames.
Moreover, 10 video-level tracking attributes are also introduced to facilitate attribute-wise tracking performance.
As compared to existing datasets, we propose a new UVOT benchmark dataset consisting of 400 sequences and 275,000 annotated bounding boxes.
The number of video-level attributes is 17 with quite diverse 50 object categories.
More details of the proposed dataset are shown in Table \ref{table1}.

\subsection{Benchmark SOTA Visual Trackers}
\label{sec:sota}
In recent years, three dominant tracking paradigms including Discriminative Correlation Filters (DCFs)-based, Siamese-based, and Transformer-based trackers have evolved and demonstrated outstanding performance on many open-air VOT benchmarks \cite{fan2019lasot, fan2021lasot, huang2019got, muller2018trackingnet}.

\textbf{DCFs-based Trackers:} In these trackers, correlation filters are trained using the region of interest in the current frame.
The learned DCF is then employed to track the target object in the subsequent frames by estimating the maximum correlation response \cite{henriques2014high}.
We benchmark four DCFs-based deep trackers including ATOM \cite{danelljan2019atom}, DiMP \cite{bhat2019learning}, PrDiMP \cite{danelljan2020probabilistic}, and ARDiMP \cite{yan2021alpha} on our proposed UVOT400 dataset.
ATOM predicts an overlap between the target object and an estimated bounding box in an end-to-end manner.
DiMP exploits both target and background appearance information for model prediction.
PrDiMP predicts the conditional probability density of the target state given an input image.
ARDiMP employed an alpha-refine module to improve the accuracy of the estimated target object bounding box.

\textbf{Siamese-based Trackers:} In these trackers, an embedding is learned while maximizing the distance between the background and target and minimizing the distance between the search region and template region. 
It learns an offline matching function in a large-scale diverse dataset.
We selected eight recent Siamese-based trackers including SiamFC \cite{bertinetto2016fully}, SiamBAN \cite{chen2022siamban}, SiamRPN \cite{li2018high}, SiamAttn \cite{yu2020deformable}, SiamGAT \cite{guo2021graph}, SiamMask \cite{wang2019fast}, SiamCAR \cite{guo2020siamcar}, and  RBO-SiamRPN++ \cite{tang2022ranking} for our UVOT400 tbenchmarking.
SiamFC introduced deep end-to-end similarity metric learning for VOT.
SiamRPN estimated the bounding box of the target object in an end-to-end manner by borrowing the object detection capabilities such as region proposal network from Faster R-CNN \cite{ren2015faster}.
SiamMask employed the segmentation mask of the target object and improved the tracking performance.
SiamCAR employed pixel-wise classification and regression using two sub-networks for target bounding box estimation to improve VOT.
SiamBAN estimated the anchor-free target bounding box in an end-to-end manner.
SiamGAT employed a graph attention network for part-to-part correspondence between the target and the search regions for improving VOT. 
SiamAttn introduced a deformable Siamese attention network consisting of deformable self-attention and cross-attention.
RBO-SiamRPN++ employed a ranking-based optimization method to estimate the relationship among different proposals. It is based on a classification-based ranking loss and an intersection over the union-guided loss.

\textbf{Transfomer-based Trackers:} Transformers have recently been used for learning discriminative embedding using a multi-head self-attention mechanism \cite{vaswani2017attention, han2022survey}.  
Transformers have been integrated with DCFs-based trackers such as TrDiMP \cite{wang2021transformer} and with Siamese-based trackers such as TrSiam \cite{wang2021transformer}.
We have selected 12 transformer-based trackers including RTS \cite{paul2022robust}, ToMP \cite{mayer2022transforming}, STMTrack \cite{fu2021stmtrack}, TransT \cite{chen2021transformer}, KeepTrack \cite{mayer2021learning}, CSWinTT \cite{song2022transformer}, TrTr \cite{zhao2021trtr}, STARK \cite{yan2021learning}, SparseTT \cite{fu2022sparsett}, and AutoMatch \cite{zhang2021learn} for benchmarking.
Trackers, TrSiam and TrDiMP,  employed a transformer-based architecture in the Siamese and DCFs-based tracking frameworks.
STMTrack employed a space-time memory network to use historical target object information to handle target appearance variations.
TrTr employed transformer-based architecture to get improved self-attention for VOT.
TransT employed an attention-based feature fusion network to combine the template and search region features.
STARK also employed transformer-based architecture that models spatiotemporal feature dependencies for VOT.
ToMP employed transformer model prediction which is then used to estimate a second set of weights for accurate bounding box regression. 
RTS tracker used segmentation masks instead of bounding boxes to reduce the background contents and improve the learning of target representation.
CSWinTT employed a transformer-based architecture with multi-scale cyclic shifting window attention for VOT.
SparseTT tracker introduced sparse attention focusing on the most relevant information in the search regions for improved VOT.
AutoMatch employed multiple matching operators for the purpose of feature fusion.
KeepTrack addressed the problem of similar distractors by tracking those distractors in addition to the target object using an association network.

Although these trackers have demonstrated excellent performance on open-air datasets we observed performance degradation across all trackers for underwater VOT.

\section{Proposed UVOT400 Dataset}
\label{sec::dataset}
The aim of this work is to propose a dedicated benchmark for the training and evaluation of underwater tracking methods.
For this purpose, we ensure that our proposed UVOT dataset is large-scale with high-quality dense annotations.
Most sequences provide long-term tracking challenges, the target objects are comprehensive, and cover a wide range of UW-moving creatures.

\subsection{Dataset Details}
Our proposed UVOT400 dataset consists of 400 sequences of UW moving targets with an average length of 688 frames per video captured at 30 \textit{fps}.
The maximum, minimum, and average frame resolutions of the sequences are $1520 \times 2704$, $959 \times 1277$, and $1080 \times 1920$.
It consists of an overall 275,358 densely annotated frames with the minimum length of a sequence being 40 frames and the maximum length being 3300 frames.  
There are 220 video sequences with fewer than 600 frames offering short-term tracking challenges while 180 sequences have larger lengths offering long-term tracking challenges.
The proposed dataset covers 17 distinct VOT attributes consisting of both UW-specific as well as common attributes of the existing open-air VOT datasets.
Our dataset contains 50 diverse target object categories of varying sizes and diverse motion patterns.
Our aim is to offer the tracking community a high-quality benchmark for underwater VOT.

\begin{figure*}[t!]
\centering
\includegraphics[width=\textwidth]{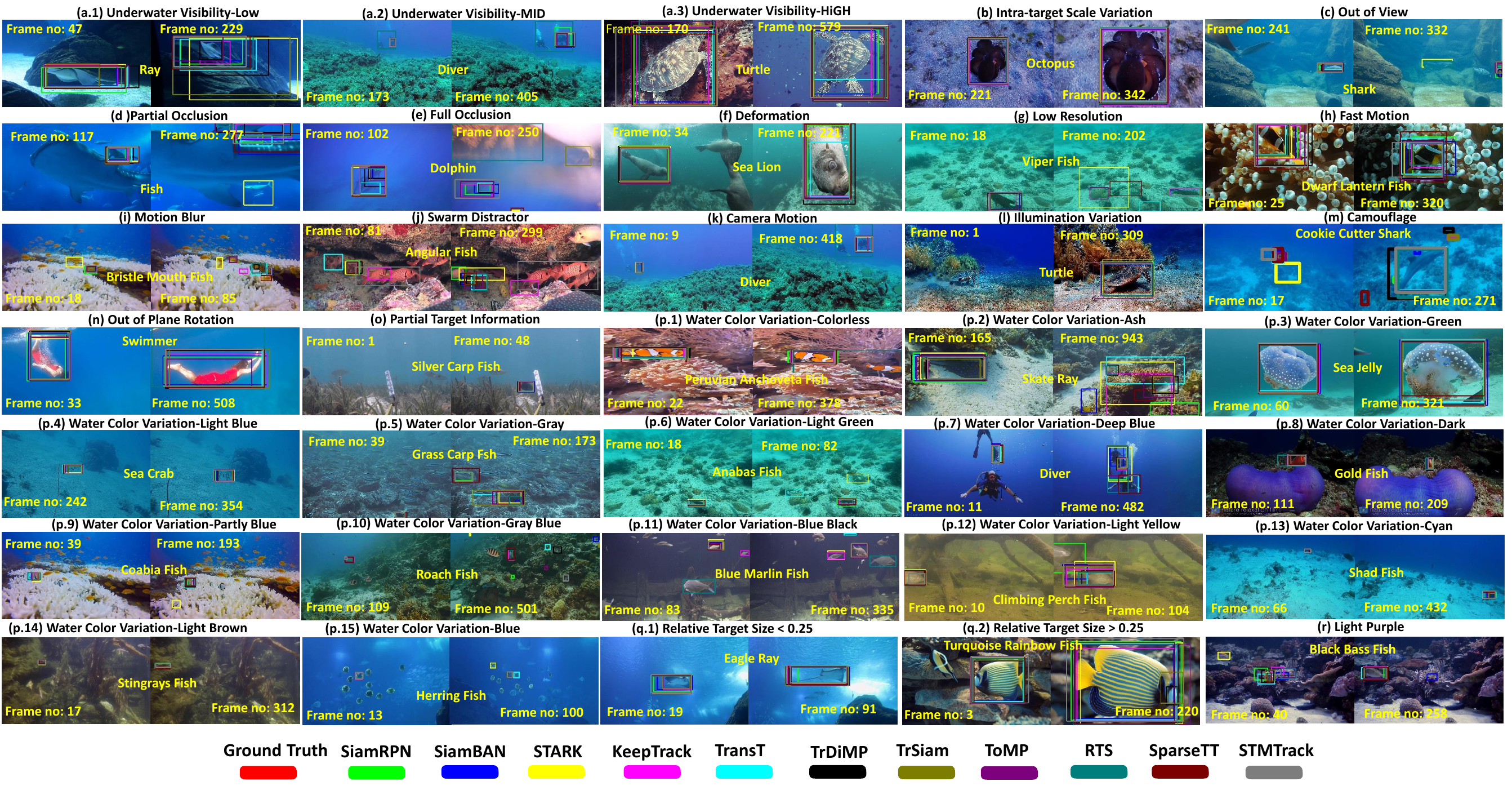}
\caption{UVOT400 dataset - sample images with targets from diverse categories
showing the tracked target object bounding boxes predicted by 12 SOTA trackers, color-coded as shown in the legend. 
(a)-(q) samples images from all 17 tracking attributes.
(a.1)-(a.3) three levels of visibility conditions -- Low, Mid, and HiGH.
(p.1)-(p.15) the 15 sub-categories of water color variation (WCV) attributes.
(q.1)-(q.2) sample images with relative target sizes less than or greater than 0.25.
}
\label{fig2}
\end{figure*}

\subsection{Dataset Collection}
UVOT400 has been obtained both locally from UAE coastal region and from publicly available online sources including YouTube, Pixabay\footnote{https://www.pexels.com/search/videos/underwater/}, and underwater change detection\footnote{http://underwaterchangedetection.eu/Videos.html}. 
The online sources have enabled us to select diverse videos for the construction of challenging benchmarks with varying backgrounds to avoid bias toward certain underwater environments.

Most videos have been captured in the wild which is quite helpful for evaluating trackers for real-world underwater applications.
Initially, a large number of videos were collected containing UW moving targets, however, in the suitability analysis, only 400 video sequences are retained.
Our suitability criteria contain the visual presence of the target object in all frames and a balance across multiple tracking attributes.
Each sequence is ensured to have a single shot because a concatenation of multiple shots introduces irregularity in the target path.
The video sequences containing irrelevant material such as personal introductions are removed.
Sequences having lengths less than 40 frames are also discarded while the sequences having lengths larger than 3300 frames are divided into multiple sequences.
Eventually, a large-scale UVOT benchmark is collected under Creative Commons license from the aforementioned resources.

\subsection{Annotation Protocol}
High-quality and accurate annotations across all sequences in a dataset are crucial for the learning of deep trackers.
Such a quality requires defining annotation protocols beforehand.
While annotating the UVOT dataset, we have also followed a fixed protocol consisting of several steps.
For the 1st frame in each sequence, an accurate axis-aligned bounding box precisely covering the target object is manually annotated.
In the remaining frames of the sequence, a rough bounding box is obtained for the annotated target object using Computer Vision Annotation Tool (CVAT) \footnote{https://www.cvat.ai/}.
Each such obtained bounding box is then manually examined by three distinct annotation teams consisting of several M.S. and Ph.D. students.
The annotations are then validated by a validation team for correctness and precision.
If an annotation is not fully containing the target object or contains more undesirable background content then it is referred back to the annotation team for corrections.
Once three corrected annotations are obtained, the final annotation is computed as an average of these annotations.
The attributes of each video sequence are also labeled by each annotation team.
The final attributes are decided by the validation team using majority voting if a conflict is found in attribute labeling.
Using this protocol, a high-quality densely annotated benchmark known as UVOT400 is obtained.

\subsection{Tracking Attributes}
In order to evaluate trackers under different types of challenges, each video sequence is labeled with 17 distinct attributes.
An attribute shows a particular video content that may pose a specific challenge to a tracker while a set of attributes may simultaneously pose a set of challenges at the same time.
Such labeling assists in a better analysis of a tracker over varying tracking conditions.
In this work, each video may contain one or more of the following 17 attributes: 

\begin{enumerate}
    \item \textbf{UW Visibility (UWV)}: UWV depends on the available lighting conditions, water quality, and distance of the target object from the camera. 
    The UVOT dataset is manually categorized by the annotators based on target object visibility into three classes including high, medium, and low visibility.
    Figs. \ref{fig2} (a.1)-(a.3) show sample images corresponding to different levels of the UW visibility conditions in the UVOT dataset.
 \item \textbf{Swarm Distractors (SD)}:  If there are many objects very similar to the target object the tracker may start tracking some other similar object. 
 In some cases, the number of these SD can be significantly higher resulting in a swarm of distractors. Sample images of this attribute are shown in Fig. \ref{fig2} (j).
 \item  \textbf{Camouflage (Cam)}: Many underwater creatures have the ability to Cam themselves to avoid adversaries. 
 Such creatures are hard to discriminate from the background resulting in the Cam challenge (Figs. \ref{fig2} (m)).
 \item  \textbf{Intra-target Scale Variation (ISV)}: As the distance of the target increases from the camera, its size may significantly decrease. It may also happen due to the pose variations of the target object. Sequences having such variations are considered as ISV tracking challenges (Figs. \ref{fig2} (b)). 
 \item \textbf{Relative Target Size variation (RTS)}: Trackers face severe challenges as the target size reduces. 
 In the UVOT400 dataset, RTS is significantly larger than in the open-air datasets. 
 We define it as the ratio of pixels covered by the target and the total frame size. 
 In our dataset, the smallest RTS is 0.001 and the larger is 0.525 (Figs. \ref{fig2} (q.1)-(q.2)). 
 \item \textbf{Water Color Variations (WCV)}: In the UVOT400 dataset, sequences are divided into the following distinct categories of UW colors including Colorless, Ash, Green, Light Blue, Gray, Light Green, Deep Blue, Dark, GrayBlue, Partly Blue, Light Yellow, Light Brown, Blue, Cyan, and Blue Black.  
 Figs. \ref{fig2} (p.1)-(p.15) \& (r) shows sample images from the sequences referring to these WCV categories.
 \item \textbf{Out of View (OV)}: If the target object completely leaves the UW scene, it will be considered as OV challenge (sample images in (Figs. \ref{fig2} (c)). 
 \item \textbf{Partial Occlusion (PO)}: The target object may observe PO in consecutive frames (Figs. \ref{fig2} (d)). 
 \item \textbf{Deformation (Def)}: Many UW target objects have the capability to deform themselves in a none rigid way posing a non-rigid Def challenge to trackers (Figs. \ref{fig2} (f)). 
 \item  \textbf{Full Occlusion (FO)}: A UW target may undergo FO by other objects for a significant amount of time (Figs. \ref{fig2} (e)).
 \item \textbf{Low Resolution (LR)}: A UW sequence may be captured in LR resulting in reduced quality. 
 Tracking objects in LR sequences is challenging due to less signal-to-noise ratio. 
 The noise may get enhanced due to the scattering of light by suspended particles (Figs. \ref{fig2} (g)). 
 \item \textbf{Fast Motion (FM)}: If the motion of the target object is larger than 20 pixels in two consecutive frames it is considered a FM object \cite{fan2021lasot} (Figs. \ref{fig2} (h)).
 \item \textbf{Motion Blur (MB)}: If the target region is blurred because of the camera motion or object motion the sequence is considered MB (Figs. \ref{fig2} (i)).
 \item \textbf{Camera Motion (CM)}: Based on camera motion, sequences in the UVOT dataset are divided into two categories including no CM and CM (Figs. \ref{fig2} (k)).
 \item \textbf{Illumination Variation (IV)}: If the illumination within the target region changes in a sequence, it is labeled as an IV sequence (Figs. \ref{fig2} (I)).
 \item \textbf{Out of Plane Rotation (OPR)}: Most of UW target rotations are out of plane. Such sequences are labeled as OPR (Figs. \ref{fig2} (n)).
  \item \textbf{Partial Target Information (PTI)}: If only a part of the target object is visible in the initial frames, it is considered as a PTI sequence (Figs. \ref{fig2} (o)).
\end{enumerate}
We observe that some of these attributes are also present in many open-air VOT datasets including OV, PO, Def, FO, LR, FM, MB, CM, IV, OPR, RTS, and ISV.
However, many attributes in UVOT400 dataset are very specific to the UW scenes such as UWV, SD, Cam, WCV, and PTI.
While most of the sequences contain more than one attribute, specifically having at least one UW-specific attribute.

\subsection{UW Target Object Categories}
The UVOT400 dataset contains nine distinct target object categories including fish, diver, octopus, turtle, shark, dolphin, swimmer, sea lion, and ray.
The fish category contains 352 different types of fishes (e.g., jellyfish, juvenile frogfish, bristle mouths, angler fish, viperfish, grass carp, Peruvian anchoveta, and silver carp, etc.), 11 types of divers, 5 types of octopus, 7 types of turtles, 5 types of sharks (e.g., dwarf lantern shark and cookie-cutter shark, etc.), 11 types of dolphins, 3 types of swimmers, 4 types of sea lions, and 2 types of rays.
Thus, the proposed dataset contains diverse categories of UW target objects.

\section{Underwater Image Enhancement}
\label{sec:UIE}
Our evaluations on several SOTA trackers using the UVOT400 dataset demonstrated significant performance drop compared to the performance of the same trackers on open-air datasets as shown in Fig. \ref{fig1}.
A major performance bottleneck is the presence of inherited UW-specific attributes such as UWV conditions and WCV etc.
We aim to reduce the performance gap between UVOT and open-air tracking by proposing a novel Underwater Image Enhancement algorithm for TRacking (UWIE-TR).
We empirically demonstrate that sequences enhanced by our proposed UWIE-TR algortihm, when employed for the training of the existing SOTA open-air trackers, significantly improves the performance.
The details of our proposed UWIE-TR algorithm are presented next.

\subsection{Proposed Underwater Image Enhancement  Algorithm for TRacking (UWIE-TR)}
In the past decade, a number of UW image enhancement methods have been proposed to improve the performance of different applications such as object detection, semantic segmentation, and VOT in underwater scenarios \cite{panetta2021comprehensive, islam2020simultaneous, anwar2020diving}.

\begin{figure}[t!]
\centering
\includegraphics[width=\linewidth]{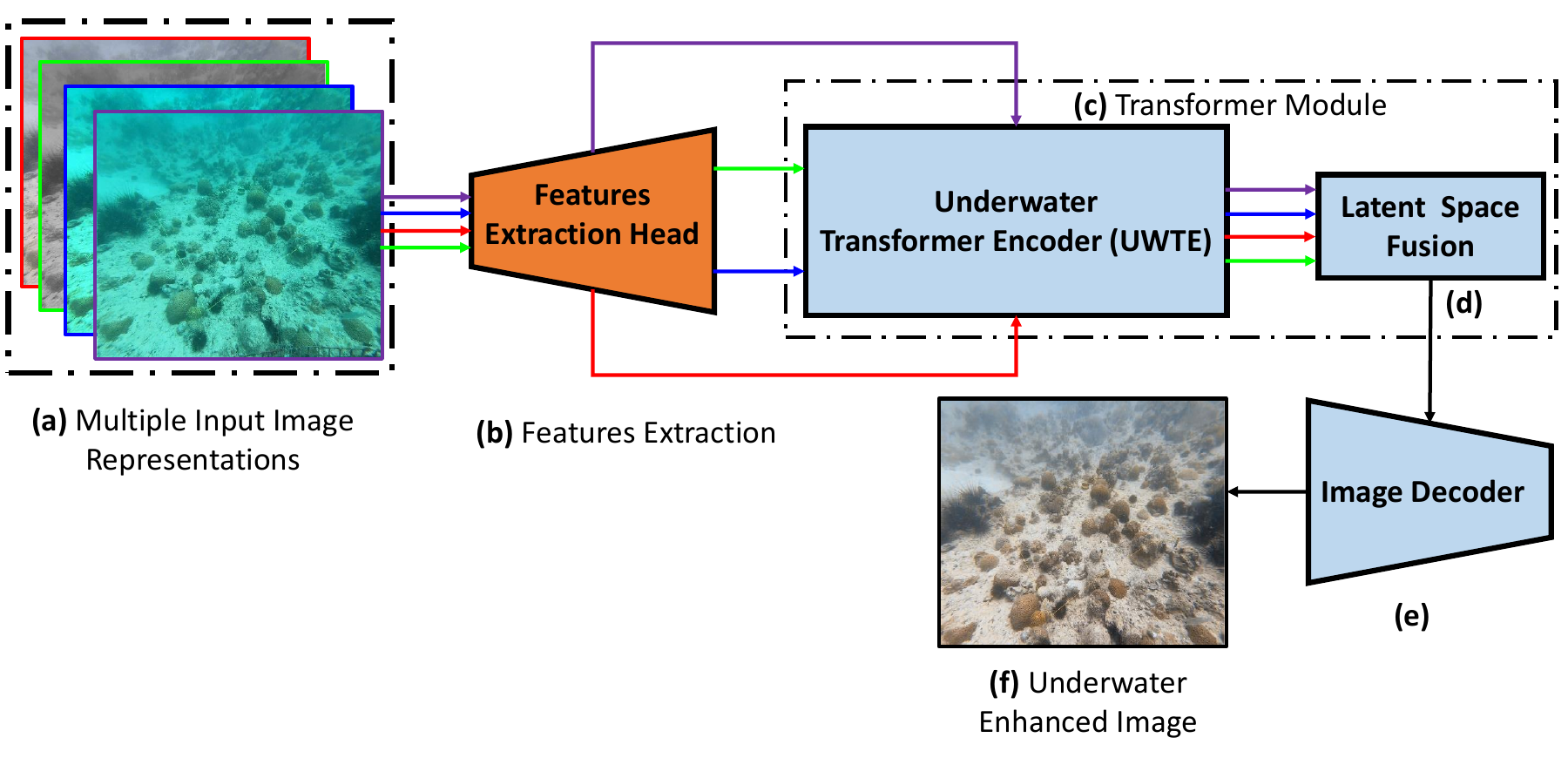}
\caption{System diagram of the proposed UWIE-TR algorithm.
(a) shows the input image and its multiple image represenations, (b) shows the feature extraction branch, (c) shows the transformer-based encoder which projects input feature representation to the latent space, (d) shows the latent space fusion using linear combination, (e) is the enhanced image decode, and (f) shows the resulting enhanced underwater image.}
\label{fig3}
\end{figure}

Transformers have recently demonstrated encouraging performance on many computer vision tasks including image denoising and restoration \cite{chen2021pre}, classification \cite{dosovitskiy2020image}, object detection \cite{carion2020end}, and VOT \cite{wang2021transformer}.
In the current work, we employ a transformer-based framework proposed by Vaswani \textit{et al.} \cite{vaswani2017attention} for UW image enhancement.
Our proposed algorithm consists of a feature extraction head, a UW transformer-based encoder, and an output image decoder as shown in Fig. \ref{fig3}.
We show that this simple architecture serves our UVOT performance improvement purpose well.

Given a UW image $\textbf{X}\in \mathbb{R}^{h \times w \times 3}$, the goal is to learn a mapping function that maps it to an enhanced target image $\textbf{Y}\in \mathbb{R}^{h \times w \times 3}$.
For each raw UW image, we generate multiple images by applying different image processing techniques such as Histogram Equalization ($\textbf{X}_{HE}$), White Balance ($\textbf{X}_{WB}$) correction, and Gamma Correction ($\textbf{X}_{GC}$) \cite{li2019underwater, ancuti2012enhancing}.
White balance ensures the white objects in the scene are rendered as white objects in the image by removing unrealistic color casts.
Many UW images appear to be bleached out or quite dark.
To correct such artifacts, we employ gamma correction which improves the overall brightness of the image.
Many UW images have their histogram contracted towards the lower intensity values resulting in low-contrast images.
In order to improve overall image quality, HE is employed to spread the histogram over the whole range of intensities.
In the below sub-sections, we explain each step of the proposed UWIE-TR algorithm in detail.

\subsubsection{Feature Extraction Head}
We employ a features extractor to extract deep image features $\textbf{F} \in \mathbb{R}^{h \times w \times c}$, where $c$ is the number of the feature maps, $h$, and $w$ denote the height and width of the input images.
Our feature extraction head consists of one convolutional layer with 64 output channels followed by two ResNet blocks each consisting of two convolutional layers each with 64 channels.
Then, we employ a transformer-based encoder architecture to learn a UW image latent space.
An image decoder is then used to generate the enhanced image.

\subsubsection{UW Transformer Encoder (UWTE)}
For input to the transformer encoder, the features \textbf{F} are reshaped to a sequence of local windows $w$ where each local window has a size of $p \times p \times c$.
The dimension of this sequence is $S\in \mathbb{R}^{p \times p \times c \times n}$, where $n$ is the total number of local windows given by $\frac{h}{p} \times \frac{w}{p}$.
To encode the position information, a learnable positional encoding $l$ is used similar to \cite{chen2021pre}.
Both positional encoding and reshaped feature sequences are input to the UWTE which consists of a Multi-head Self-Attention (MSA) layer followed by a multi-layer fully connected network.
Each UW input image representation is re-arranged as a sequence of position-aware word representation, $g=w+l$ which is input to the UWTE.
The UWTE transforms it to a learnable latent space such that $q$ be the latent representation of $g$.
Specifically, the input to the UWTE is $p_{0}=[g_{1}, g_{2},..., g_{n}]$ and the subsequent encoder steps are as follows:

\begin{equation}
\begin{split}
   & q_{x}=k_{x}=v_{x}=\textbf{LN}(p_{x-1}),~\hat{p}_{x}=\textbf{MSA}(q_{x},k_{x},v_{x})+p_{x-1}, \\
    &p_{x}=\textbf{MLP}(\textbf{LN}( \hat{p}_{x}))+\hat{p}_{x}, \\
    &p_{L}=[q_{1}, q_{2},..., q_{n}],
    \end{split}
\label{eqn2}
\end{equation}

\noindent where $x=1,2,...,L$ denotes the number of different UWTE layers and $LN$ represents the Layer Normalization \cite{ba2016layer}.

\begin{figure*}[t!]
\centering
\includegraphics[width=\linewidth]{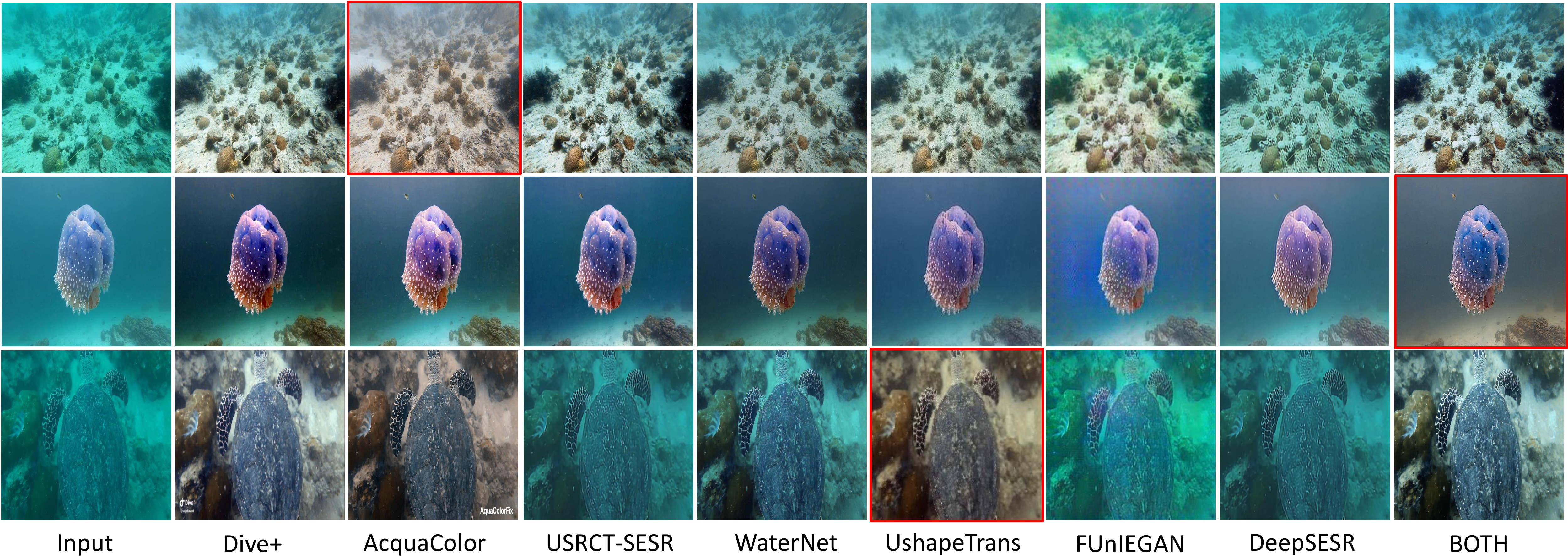}
\caption{Image enhancement results of published methods. From left to right: raw underwater images, Dive+, AcquaColor, USRCT-SESR \cite{ren2022reinforced}, WaterNet \cite{li2019underwater}, UshapeTrans \cite{peng2023u}, FUnIEGAN \cite{islam2020fast}, DeepSESR \cite{islam2020simultaneous}, and BOTH \cite{liu2022boths}. Red boxes -- the 
images picked by the human experts as pseudo groundtruth .}
\label{fig4}
\end{figure*}

\subsubsection{Latent Space Fusion}
We fuse the latent space representation of the different pre-processed versions of the UW images using a linear combination.
Each of the input image \textbf{X} and its different pre-processed versions such as $\textbf{X}_{WB}$, $\textbf{X}_{GB}$, and $\textbf{X}_{HE}$ are input to features extraction head to get different feature representations including $\textbf{F}$, $\textbf{F}_{WB}$, $\textbf{F}_{GB}$, and $\textbf{F}_{HE}$.
For each feature representation of these images, we get latent space representations using the UWTE as $\textbf{U}$, $\textbf{U}_{WB}$, $\textbf{U}_{GB}$, and $\textbf{U}_{HE}$.
The latent space representations are fused using a linear combination as follows:

\begin{equation}
\widehat{\textbf{U}}=\textbf{U}+\alpha\textbf{U}_{WB}+\beta \textbf{U}_{GB}+ \gamma\textbf{U}_{HE},
\label{eqn2}
\end{equation}

\noindent where $\alpha$, $\beta$, and $\gamma$ are learnable hyper-parameters.

\subsubsection{Enhanced Image Decoder}
The fused representation obtained in the latent space consists of 64 channels.
It is then decoded using an enhanced image decoder consisting of a convolutional layer followed by two ResNet blocks each consisting of two convolutional layers with 64 channels with $3 \times 3$ kernel size and $3$ output channels. 
The complete architecture is trained in an end-to-end manner using $\ell_{1}$ appearance loss, $\ell_{1}$ perceptual loss, and $\ell_{1}$ latent space loss function between the generated and the ground truth images as:

\begin{equation}
\mathbfcal{L}=\lambda_{1}|| \textbf{Y}  -\widehat{\textbf{Y}}||_{1} + \lambda_{2}||\textbf{F}_{\textbf{Y}} -\textbf{F}_{\widehat{\textbf{Y}}}||_{1} + \lambda_{3}|| \textbf{U}_{\textbf{Y}}  -\textbf{U}_{\widehat{\textbf{Y}}}||_{1},
\label{eqn1}
\end{equation}

\noindent where \textbf{Y} is the pseudo-groundtruth generated image and $\widehat{\textbf{Y}}$ is the enhanced UW image, $\textbf{F}_{\textbf{Y}}$ and $\textbf{F}_{\widehat{\textbf{Y}}}$ are the deep feature representations of pseudo-groundtruth and enhanced images computed using features extraction head, and $\textbf{U}_{\textbf{Y}}$ and $\textbf{U}_{\widehat{\textbf{Y}}}$ are different latent space representation of pseudo-groundtruth and enhanced images computed using UWTE. 
The hyper-parameters $\lambda_{1}$, $\lambda_{2}$, and $\lambda_{3}$ give a relative importance to the three loss terms.

\subsection{Underwater Image Enhancement Evaluations}
In this section, we evaluate the performance of the enhanced images estimated by our proposed UW image enhancement algorithm in detail.
We first discuss the training details of the proposed algorithm followed by performance comparison with SOTA underwater image enhancement methods.

\subsubsection{Pseudo-GroundTruth Generation}
The proposed UVOT400 dataset does not contain any reference images as groundtruth data, therefore, we have to employ the existing SOTA UWIE methods to generate pseudo-groundtruth images similar to Li \textit{et al.} \cite{li2019underwater}. 
For this purpose, existing methods are employed including Fusion \cite{ancuti2012enhancing}, Two Step \cite{fu2017two}, Retinex \cite{fu2014retinex}, Deep WaveNet \cite{sharma2023wavelength}, UDCP \cite{drews2016underwater}, Dive+ \footnote{
\label{d} http://dive.plus/}, AcquaColor \footnote{\label{e} https://www.aquacolorfix.com/}, USRCT-SESR \cite{ren2022reinforced}, WaterNet \cite{li2019underwater}, Ushape Transformer (UshapeTrans) \cite{peng2023u}, FUnIEGAN \cite{islam2020fast}, DeepSESR \cite{islam2020simultaneous}, and BOTH \cite{liu2022boths}.

Since the proposed UVOT400 is a video-based dataset, therefore each video is divided into 10 segments of equal length, and one random frame is selected from each segment.
Thus, 400 sequences result in 4,000 images which are enhanced using each of the above-mentioned methods and named as UW4K dataset.
The generated pseudo-groundtruth images are then manually evaluated by a team of 10 experts.
Each expert selected the best-enhanced image for each raw video frame.
Based on the majority voting, the best-enhanced image is selected.
Since, from each video, we obtain 10 frames, therefore, majority voting is also performed on each video and a single enhancement method is selected.
Fig. \ref{fig4} shows the enhanced UW images generated by the eight SOTA methods.
The selected UW-enhanced images are used as a pseudo-ground truth in the training of our proposed UWIE-TR algorithm.
The full video sequence is then enhanced using the selected enhancement method and employed for the training of SOTA trackers.

\subsubsection{Training Details}
Our proposed UWIE-TR is a fully supervised learning algorithm requiring pseudo-groundtruth images to be used as targets during training.
The proposed UW4K is then divided into training and testing sets as 70$\%$-30$\%$.
The training set contains 2800 images which are not enough to train the proposed architecture.
Therefore, we have to employ the training splits of the external datasets including UFO-120 (1500 training images) \cite{islam2020simultaneous}, EUVP (11435 images) \cite{islam2020fast}, HICRD (1700 images) \cite{han2022underwater}, and UIEB (890 training images) \cite{li2019underwater}.
These datasets have often been used for training and testing the UWIE methods.
Overall, our training data contains 18,325 UW images with reference images.
\textit{Please see our supplementary material for more details on the performance evaluations of the proposed UWIE algorithm.}

\section{UVOT400 Dataset Evaluations}
\label{sec:results}
The details of the selected SOTA trackers are given in Sec. \ref{sec:sota}.
Here, we discuss the details of the evaluation protocols, metrics, and experimental settings.
\subsection{Evaluation Protocols}
We evaluate the proposed dataset using three distinct protocols including generalization, finetuning on raw UVOT400, and finetuning on the enhanced UVOT400 dataset.
In the first protocol (\textbf{Protocol I}), we evaluated the generalization capability of different trackers trained on existing open-air datasets to our UVOT400 dataset.
For this purpose, pre-trained trackers are tested on the complete UVOT400 dataset.
For this protocol, comprehensive results are reported for the complete dataset as well as to make comparison easier, results are also reported on the testing split of the proposed dataset.

In the second protocol (\textbf{Protocol II}), different trackers are fine-tuned on the training split of the proposed UVOT400 dataset and tested on the pre-defined testing split.
The dataset is divided into training/testing splits using the 70/30 rule on the number of sequences such that all object categories and all attributes appeared in both splits.
The training sequences contain 280 videos while the testing split contains 120 sequences.
Also, no overlap is allowed between training and testing splits.

In the third protocol (\textbf{Protocol III}), different trackers are fine-tuned on the enhanced UVOT400 dataset and tested on the enhanced testing set.

\subsection {Evaluation Metrics}
We evaluate the performance of the different trackers on our proposed UVOT400 dataset using precision, normalized precision, and success rate for all protocols using One Pass Evaluation (OPE).
Most open-air trackers also employ the same VOT performance measures.

The precision is measured as the number of frames having Euclidean distance between the centers of the ground truth bounding box and the tracking result is less than the pre-defined threshold mostly 20 pixels.
The precision plot is then generated by varying the threshold from 0 to 50 pixels and average precision is reported.
The precision is often reported however it does not cater scale of the ground truth bounding box.
Therefore, normalized precision has also been proposed which incorporates the diagonal of the target object bounding box \cite{fan2021lasot}.

The success rate (SC) is computed as the ratio of the number of successfully tracked frames (i.e., intersection-over-union (IoU) between ground truth bounding box and tracking result
larger than a pre-defined threshold, typically, 0.5 to the number of all frames in a sequence.

\subsection{Implementation Details}
For protocol \textbf{I}, pre-trained trackers on the open-air datasets are employed for the UVOT400 dataset.
For each tracker, the default parameters as recommended by the original authors are used.
For protocol \textbf{II}, selected trackers are fine-tuned on the raw underwater videos of the training split of the UVOT400 dataset while reducing the learning rate by four times as used in the original training.
The inference is performed using the raw videos of the testing split of the proposed dataset.
For protocol \textbf{III}, the same selected trackers are finetuned and evaluated on the training and testing splits of enhanced UW sequences of the proposed dataset.
The other parameters of the selected SOTA trackers are kept the same as in protocol \textbf{II}.

All experiments are conducted on a workstation with a 128 GB of memory, CPU Intel Xeon E5-2698 V4 2.2 Gz (20-cores), and two Tesla V100 GPUs. 
All the trackers are implemented using the official source codes provided by the respective authors.
The parameters $\lambda_{1}$, $\lambda_{2}$, and $\lambda_{3}$  for training the proposed UWIE-TR architecture are selected as the inverse of the feature volume. 
Fusion parameters $\alpha$, $\beta$, and  $\gamma$ in Eq. (\ref{eqn2}) are learnt during training.

\def \widthforthreecolumns {0.48\columnwidth}
\def \heightforvspace {0.025in}
\begin{figure*}[t]
\centering
\subfigure{\begin{minipage}{\widthforthreecolumns}
\includegraphics[width=\textwidth,height=\linewidth]{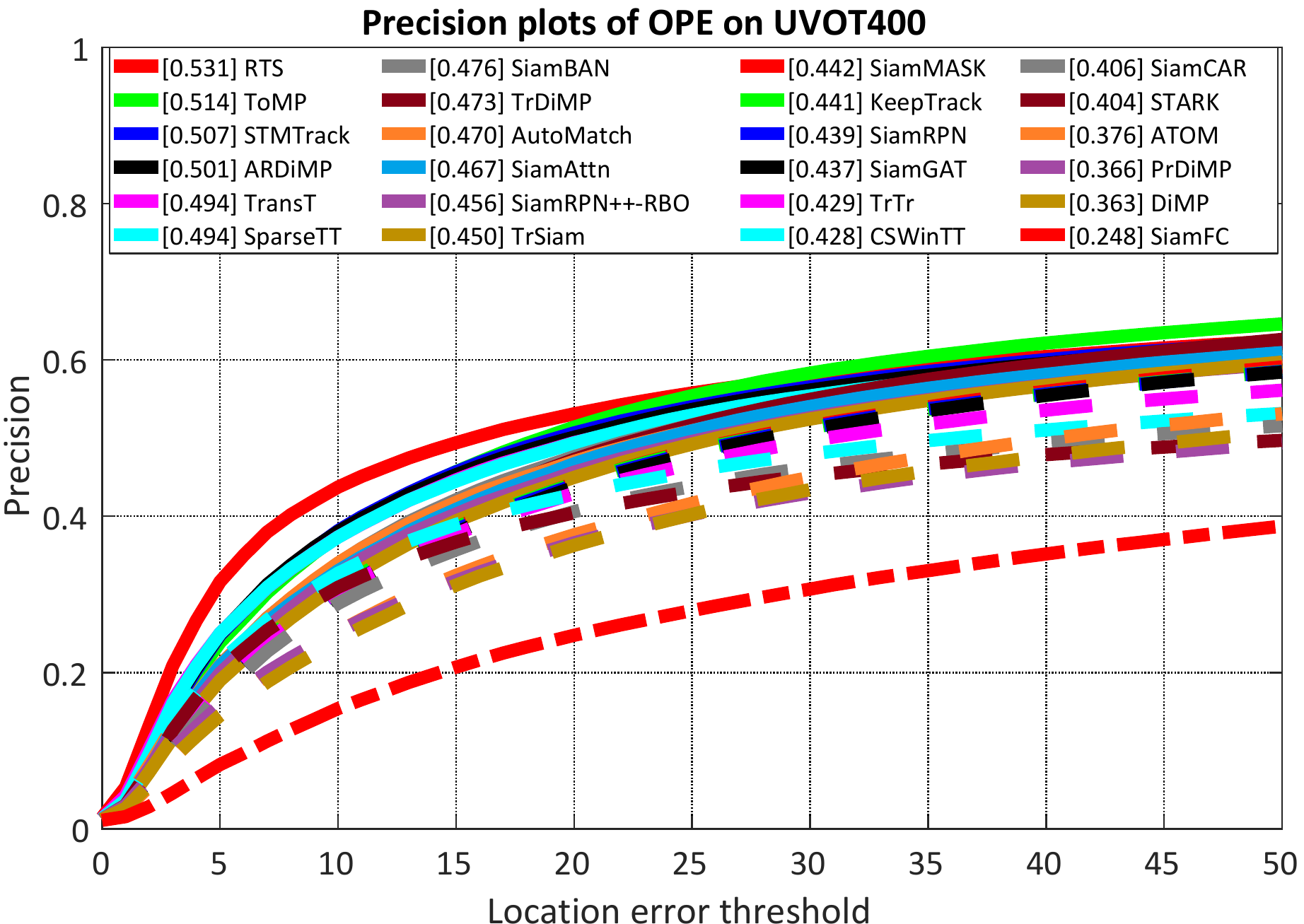}\caption*{(a) Precision Plot}\vspace{\heightforvspace}
\includegraphics[width=\textwidth,height=\linewidth]{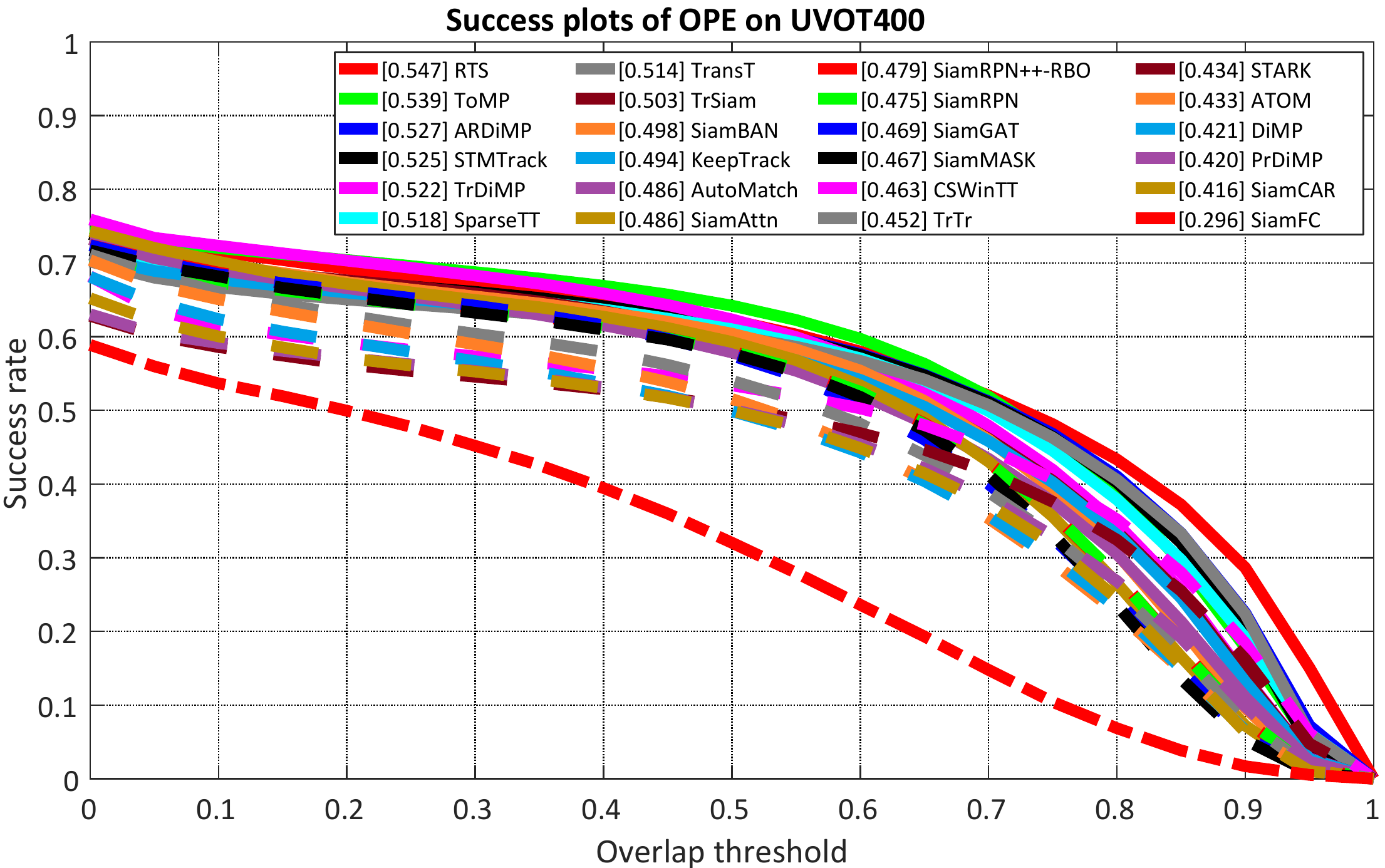}\caption*{(b) Success Plot}\vspace{\heightforvspace}
\includegraphics[width=\textwidth,height=\linewidth]{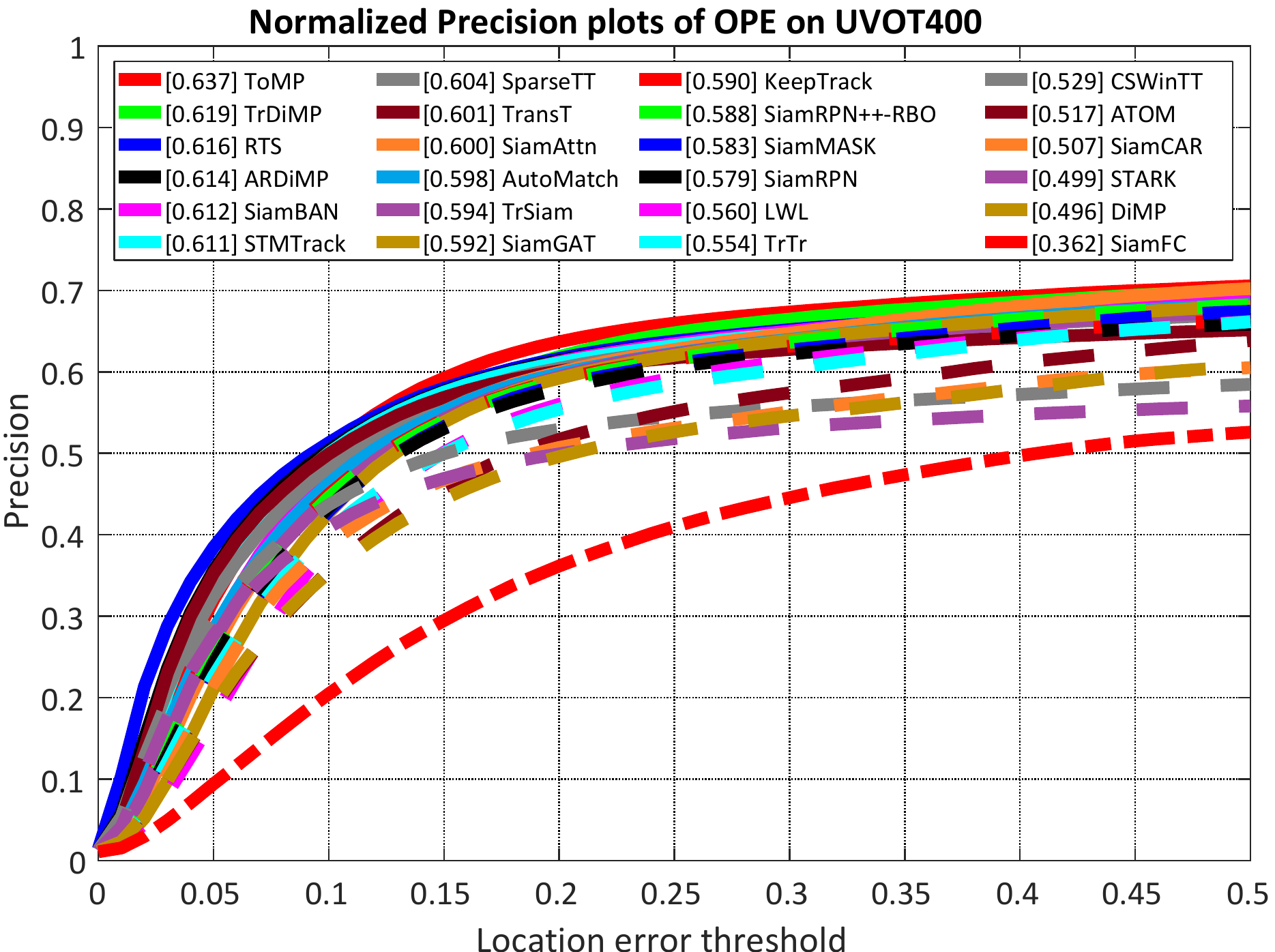}\caption*{(c) Normalized Precision Plot}\vspace{\heightforvspace}
\end{minipage}} 
\subfigure{\begin{minipage}{\widthforthreecolumns}
\includegraphics[width=\textwidth,height=\linewidth]{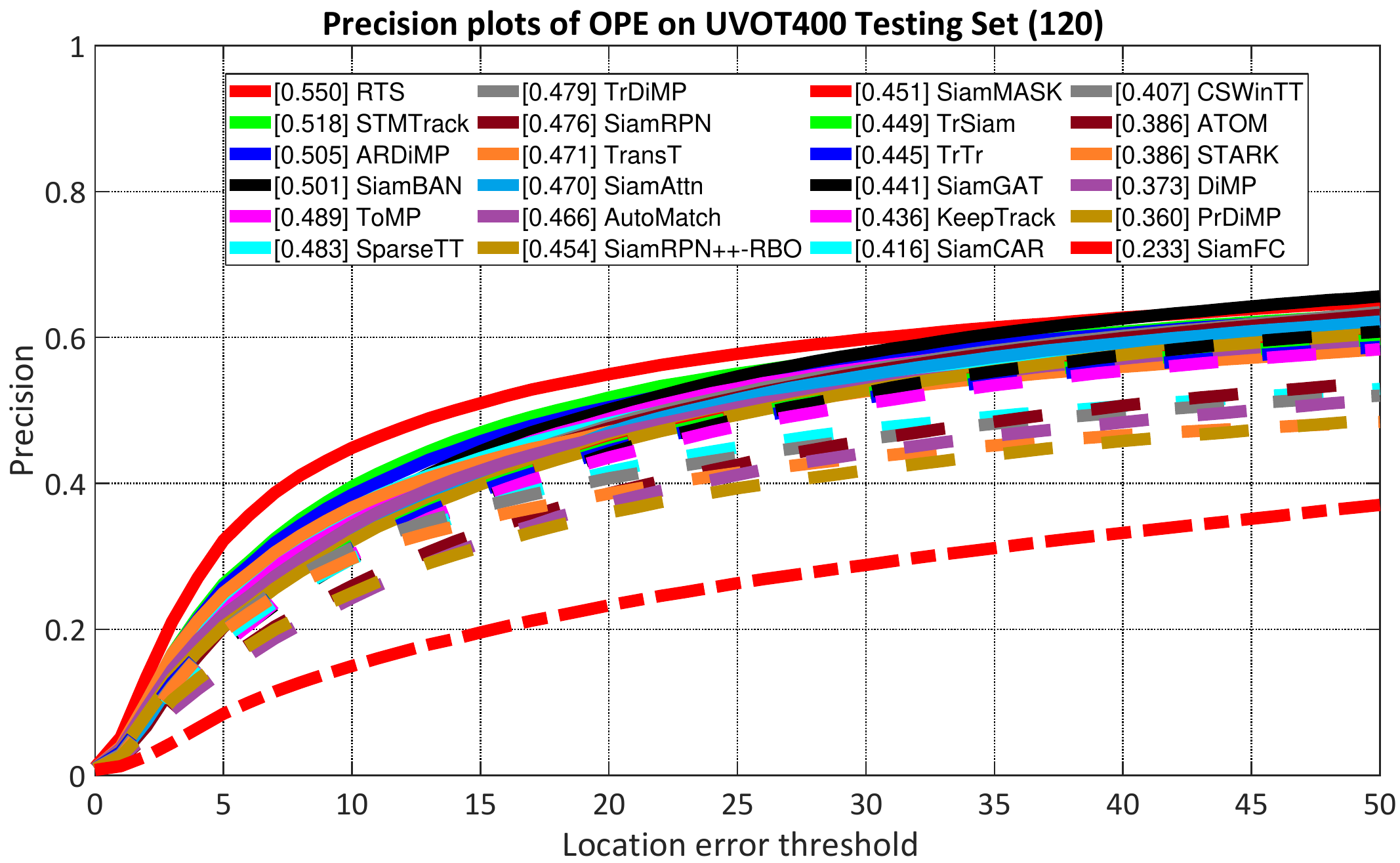}\caption*{(d) Precision Plot}\vspace{\heightforvspace}
\includegraphics[width=\textwidth,height=\linewidth]{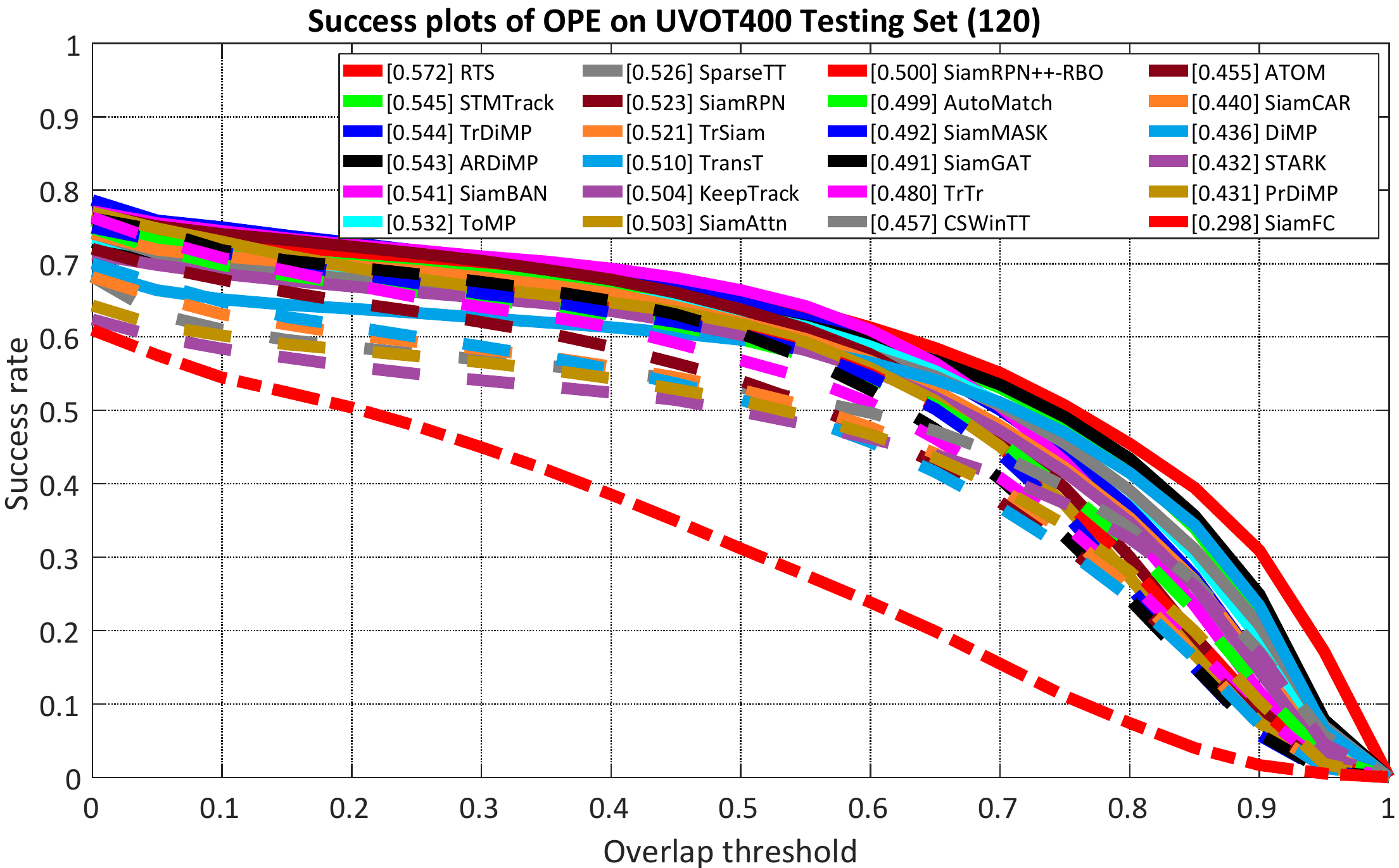}\caption*{(e) Success Plot}\vspace{\heightforvspace}
\includegraphics[width=\textwidth,height=\linewidth]{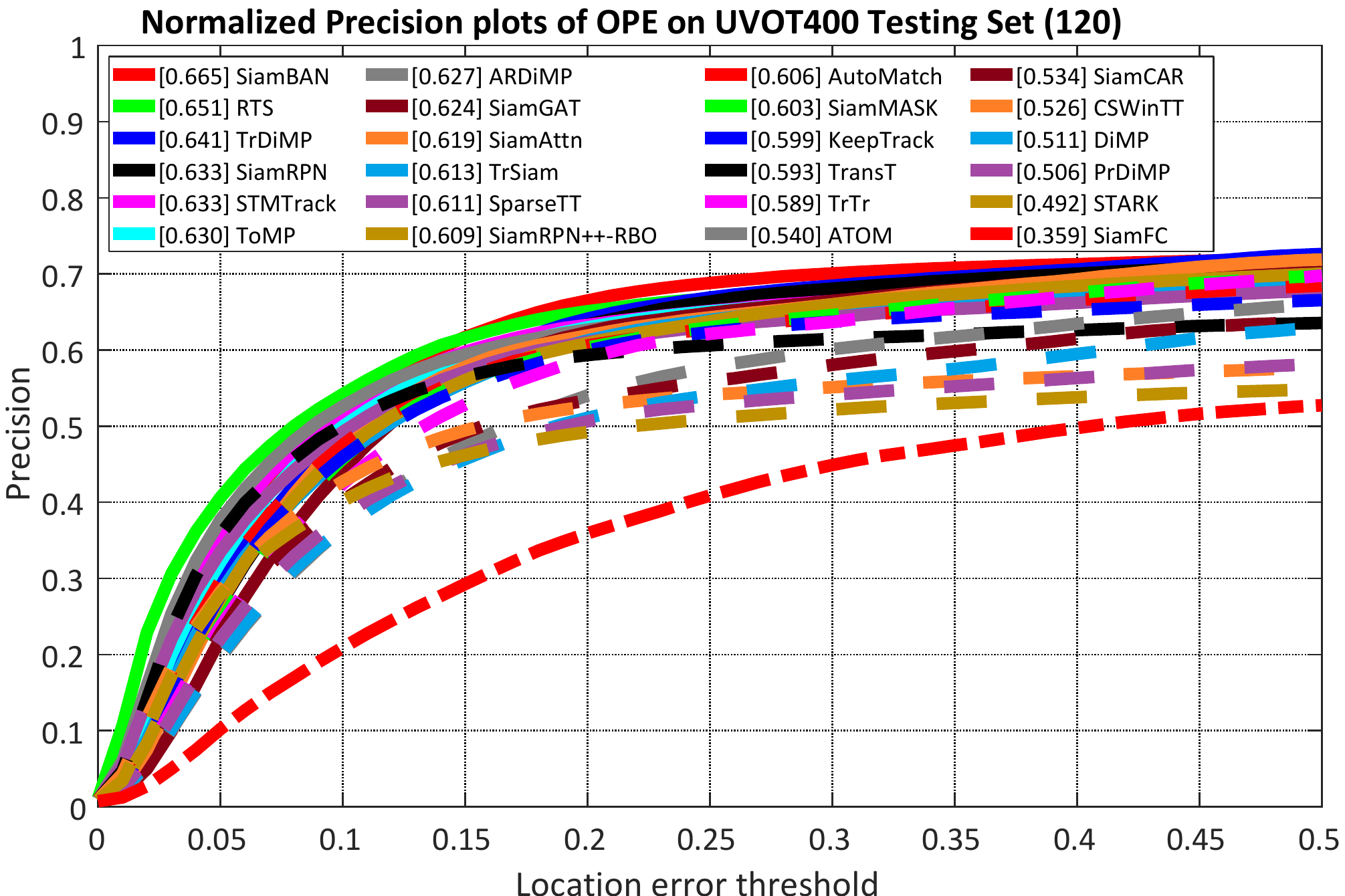}\caption*{(f) Normalized Precision Plot}\vspace{\heightforvspace}
\end{minipage}}  
\subfigure{\begin{minipage}{\widthforthreecolumns}
\includegraphics[width=\textwidth,height=\linewidth]{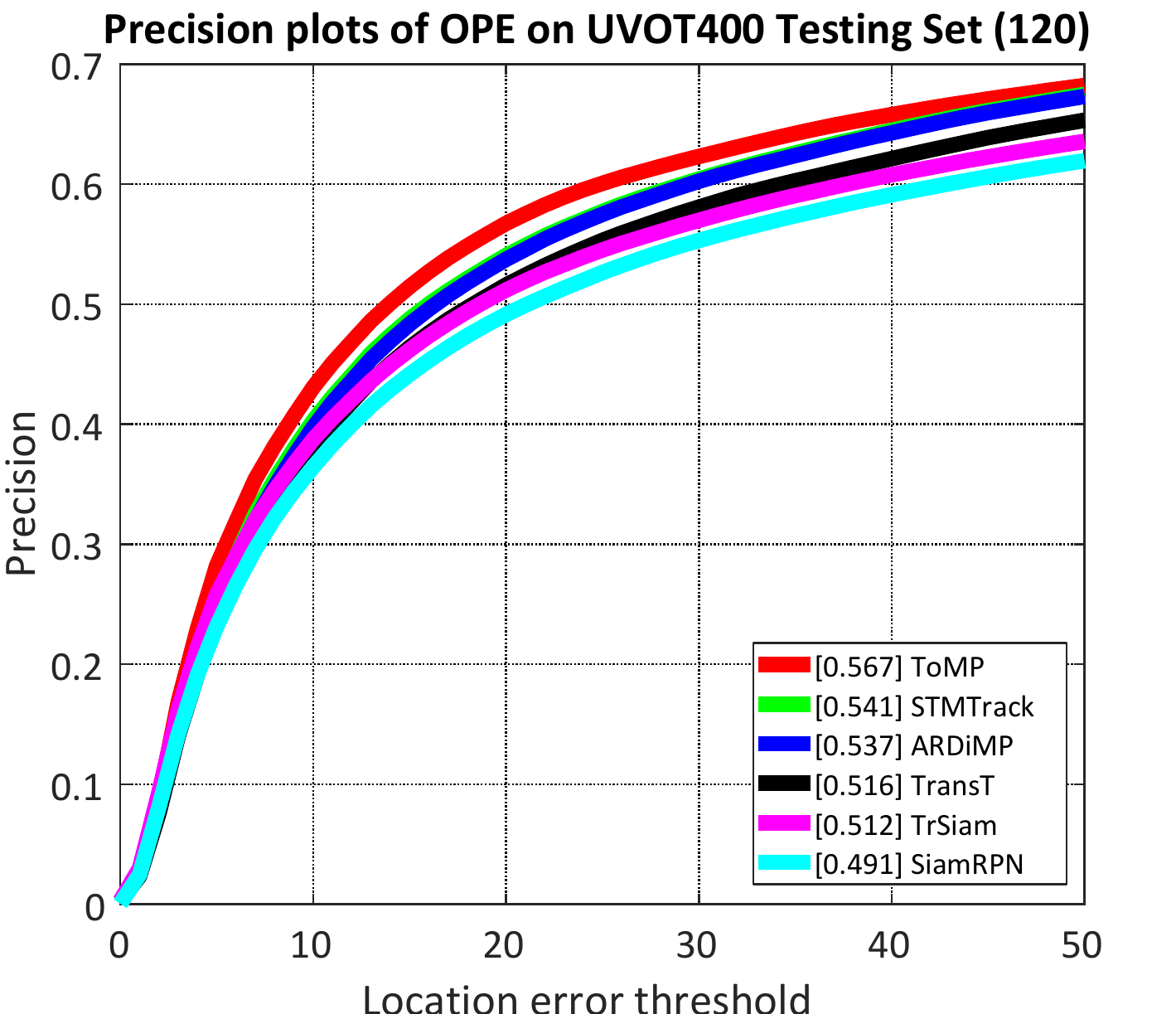}\caption*{(g) Precision Plot}\vspace{\heightforvspace}
\includegraphics[width=\textwidth,height=\linewidth]{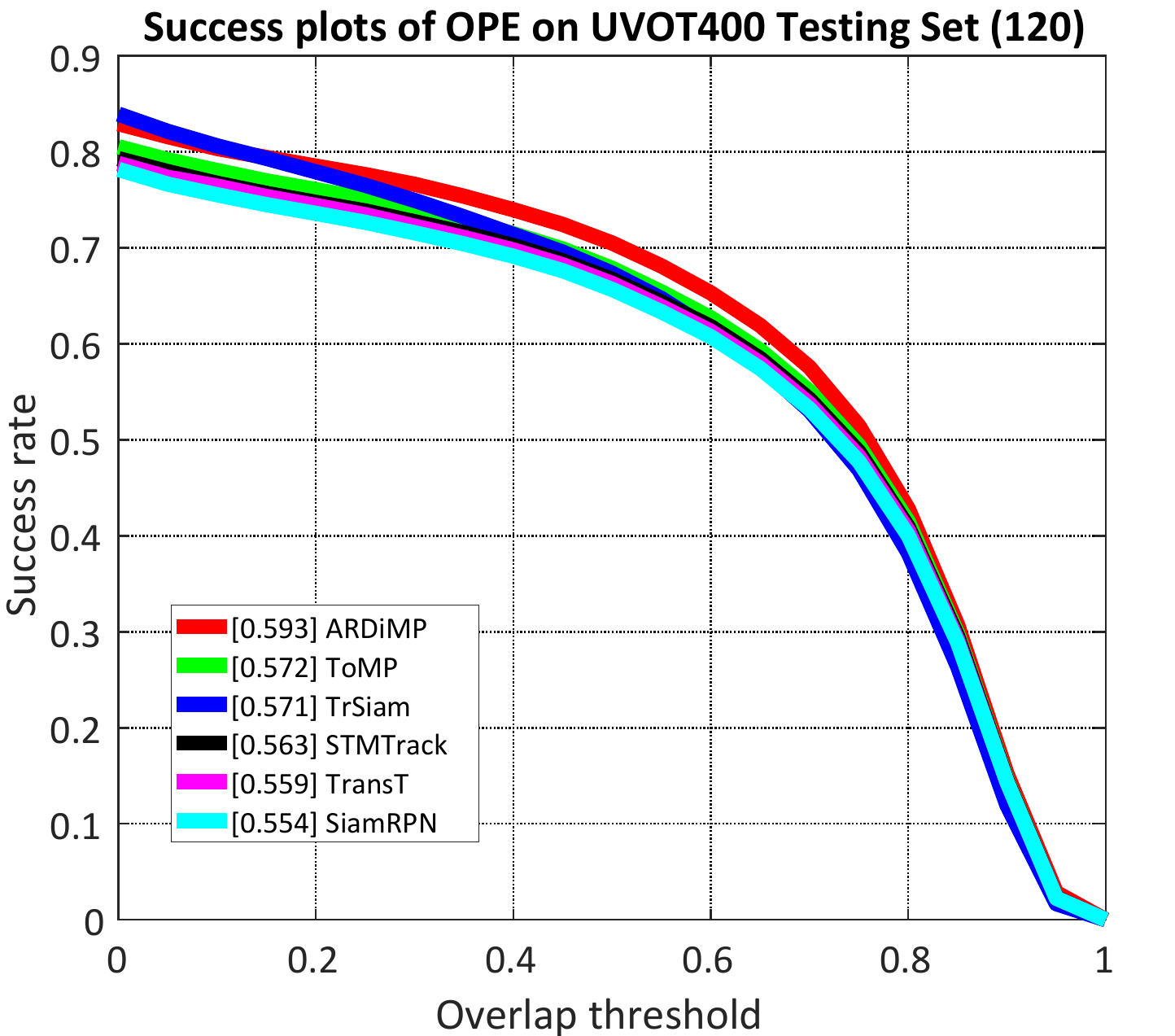}\caption*{(h) Success Plot}\vspace{\heightforvspace}
\includegraphics[width=\textwidth,height=\linewidth]{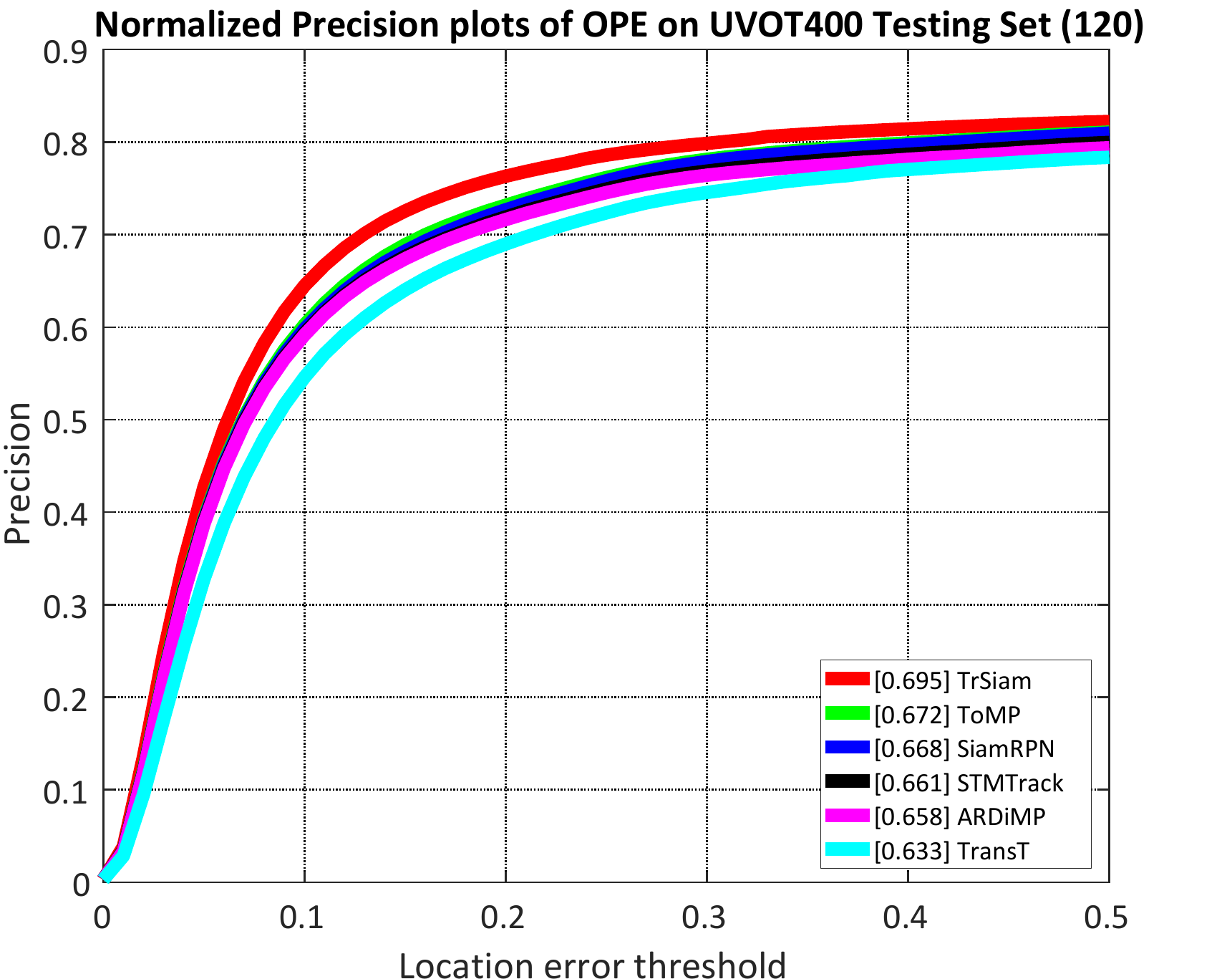}\caption*{(i) Normalized Precision Plot}\vspace{\heightforvspace}
\end{minipage}}  
\subfigure{\begin{minipage}{\widthforthreecolumns}
\includegraphics[width=\textwidth,height=\linewidth]{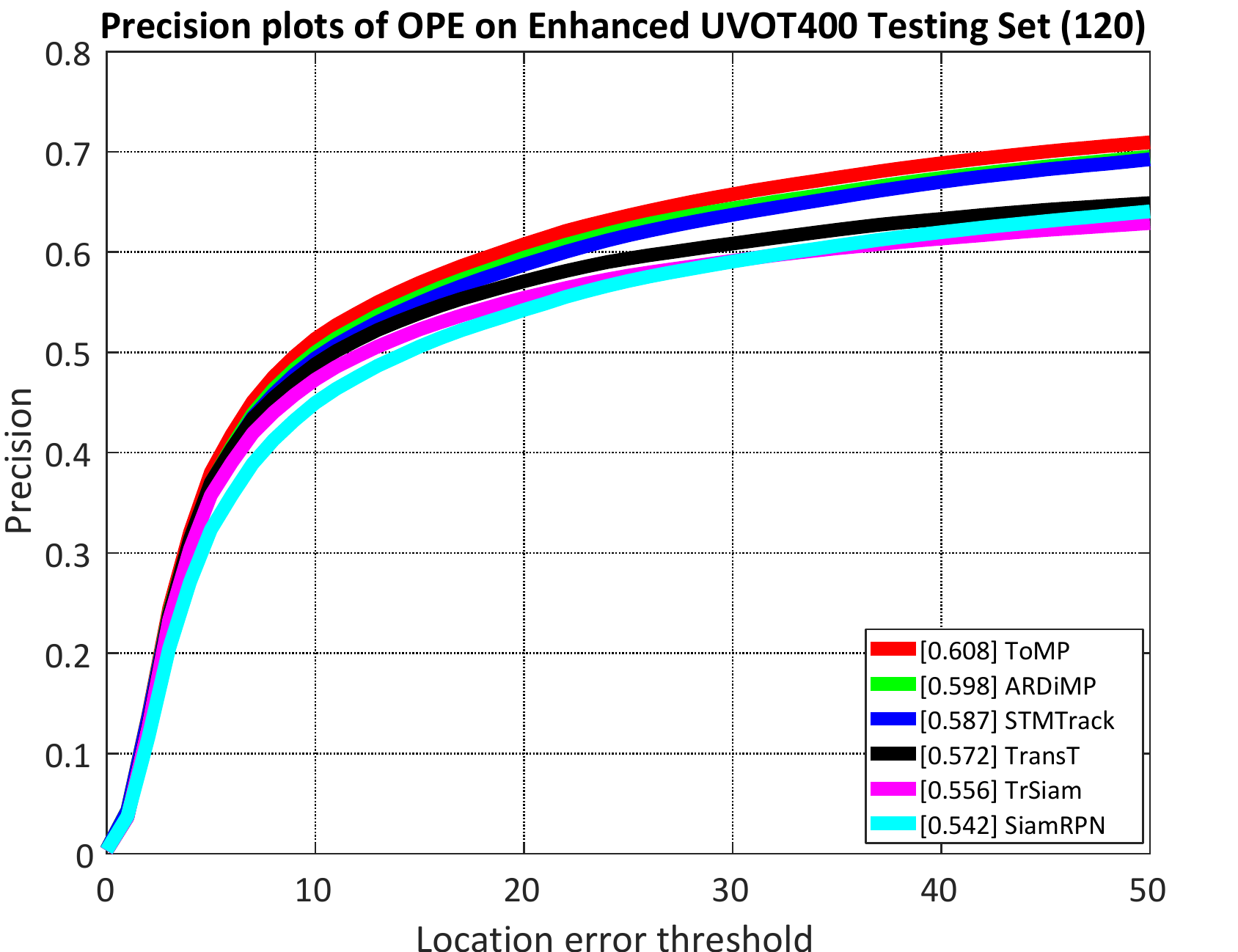}\caption*{(j) Precision Plot}\vspace{\heightforvspace}
\includegraphics[width=\textwidth,height=\linewidth]{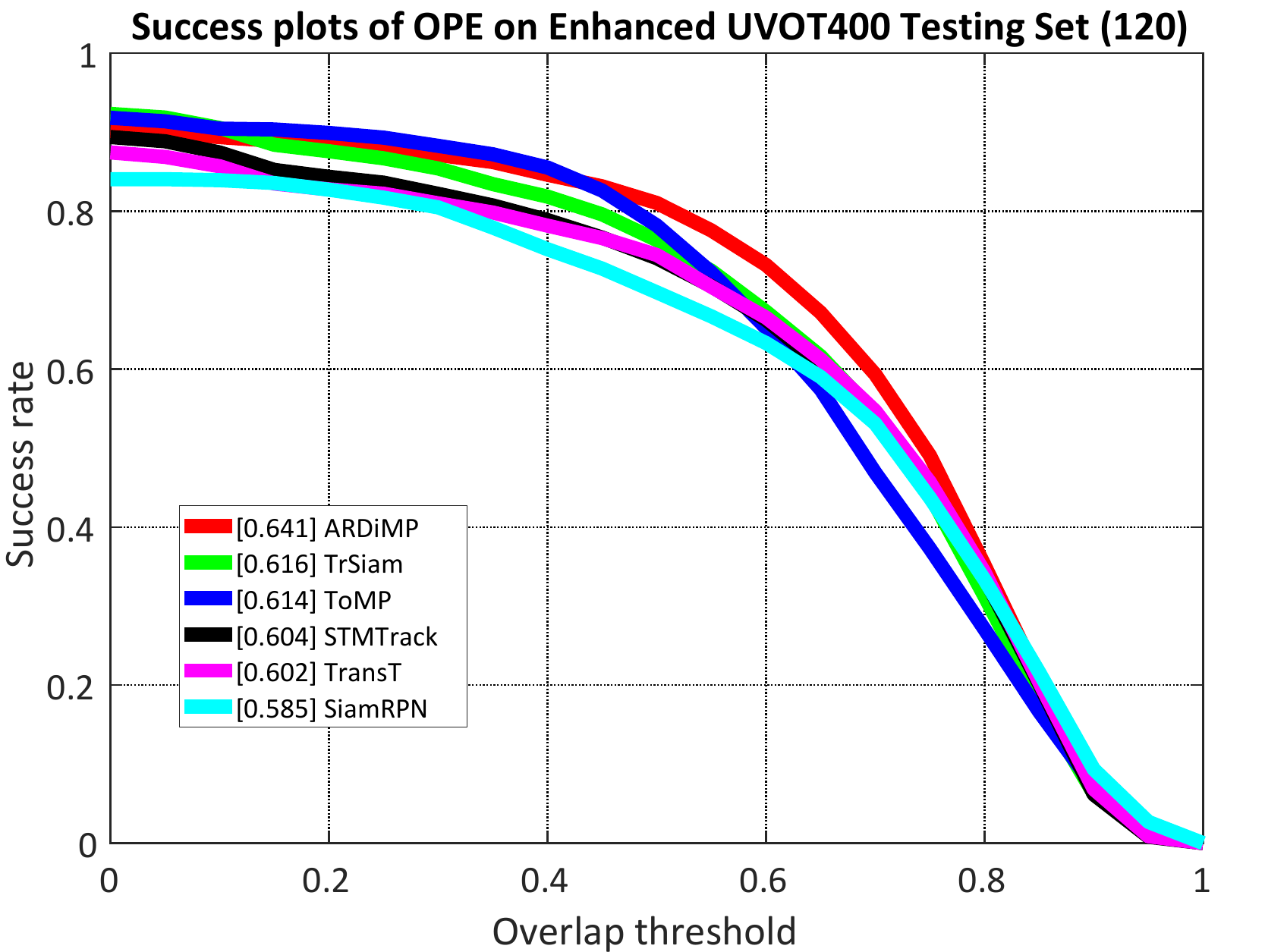}\caption*{(k) Success Plot}\vspace{\heightforvspace}
\includegraphics[width=\textwidth,height=\linewidth]{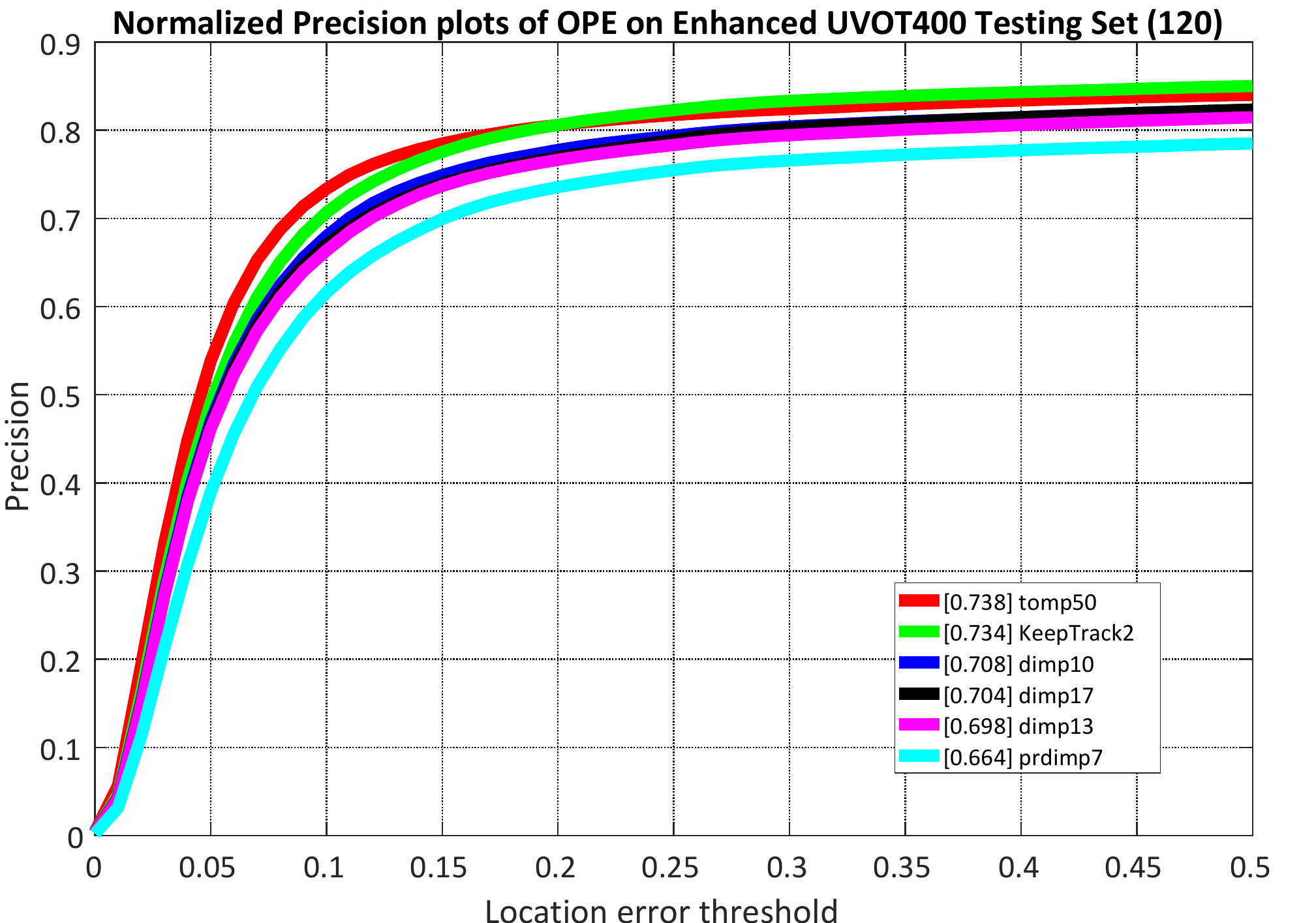}\caption*{(l) Normalized Precision Plot}\vspace{\heightforvspace}
\end{minipage}} 
\caption{\textbf{Evaluations using Protocol I, II, and III.} Protocol I: (a)-(c) -- the overall evaluation results on UVOT 400 using pre-trained trackers in terms of the precision, success, and normalized precision plots using OPE on 24 trackers.
(d)-(f) the precision, success, and normalized precision plots of the SOTA trackers using the testing split of the proposed dataset.
Protocol II: (g)-(i) the performance of fine-tuned trackers on the raw dataset.
Protocol III: (j)-(l) -- the performance of the fine-tuned trackers on the enhanced version of our proposed dataset.
The legend of the precision plot contains threshold scores at 20 pixels, while the legends of normalized precision and success plots contain area-under-the-curve scores for each tracker. 
Best viewed in color.
}
\label{fig5}
\end{figure*}

\subsection{Evaluation using Protocol I}
In this protocol, pre-trained SOTA trackers on the open-air datasets are evaluated on the proposed UVOT400 dataset.
We evaluate SOTA trackers on the overall sequences as well as attributes-based.
This protocol aims to provide large-scale evaluations of the trackers.

\subsubsection{Overall Performance}
\textbf{Complete Datset Evaluations}: The overall performance in terms of precision, success, and normalized precision plots of the evaluated trackers is shown in Figs. \ref{fig5} (a)-(c).
We observe that among the evaluated trackers on the complete dataset, RTS has obtained the best success rate of 54.70$\%$ and the best precision of 53.10$\%$ while in terms of normalized precision, the ToMP tracker achieved the best performance of 63.70$\%$.

Among the 24 evaluated trackers, only four trackers including RTS, ToMP, STMTrack, and ArDiMP have obtained both success and precision rates of more than 50.00$\%$.
In terms of normalized precision, these four trackers have obtained more than 60.00$\%$ performance.
These four trackers are based on powerful transformer architecture which is leveraged to obtain good performance on the challenging UVOT400.
Overall, most of the trackers observed performance drop compared to the open-air results as shown in Fig. \ref{fig1}.
This experiment demonstrates the challenging nature of the inherited UW attributes such as UWV and WCV etc. 

The three best trackers including RTS, ToMP, and STMTrack have observed an average performance drop of 27.43$\%$ in terms of success rate compared to the TrackingNet open-air dataset performance.
This performance drop highlights the need for developing robust underwater trackers.\\

\noindent \textbf{Testing Dataset Evaluation}: The performance comparison of SOTA trackers on the testing split of the proposed dataset is shown in Figs. \ref{fig5} (d)-(f), respectively.
In the testing set, RTS has remained the best tracker in terms of precision and success measures while SiamBAN has achieved the best performance of 66.50$\%$ in terms of normalized precision.

\subsubsection{Attribute-wise Performance}
\textit{Please see our supplementary material for detailed attributes-wise performance comparisons. }
Overall, in various UWV conditions, transformer-based trackers including RTS, STMTrack, and ToMP have performed better.
In medium UWV conditions, Siamese trackers such as SiamATTn and SiamBAN show good results while in high UWV conditions, DCFs-based trackers ARDiMP have also exhibited good performance.
Two recent trackers RTS and ToMP have performed best in almost all tracking attributes except FO in which KeepTrack remains best and LR in which STMTrack performed better.
STMTrack tracker has also performed best in SD, CAM, OPR, and PTI tracking challenges.
TrDiMP has shown good performance in IV and PTI challenges while SiamAttn has shown in SD and SiamBAN in MB attributes.
ARDiMP, SparseTT, and TransT have demonstrated good performance in IV, CAM, and OV challenges.
Overall, we observed that the transformer-based trackers performed better than the Siamese and DCFs-driven trackers.
For sequences undergoing distinct WCV attributes, the gray attribute posed the greatest challenge to the SOTA trackers in terms of precision while the Light Blue attribute demonstrated significant performance degradation in terms of success rate and normalized precision.
Overall, ToMP and STMTrack remained the best performing trackers in WCV attribute other trackers including RTS, ARDiMP, and SparseTT also performed quite well.

\begin{table*}[t!]
\caption{\textbf{Protocol I}: AUC on UVOT400 of SOTA trackers, after the application of published underwater image enhancement and the proposed UWIE-TR methods (right column). For all tested trackers, UWIE-TR has the largest performance gain (marked red; second best in blue).}
\begin{center}
\makebox[\linewidth]{
\begin{tabu}{|[2.0pt]c|c|c|c|c|c|c|c|c|[2.0pt]}
\tabucline[2.0pt]{-}
Trackers&RawImages&WaterNet&FUnIEGAN&TUDA&UColor &Dive+&AcquaColor&Prop. UWIE-TR\\\tabucline[0.5pt]{-}
ToMP&0.532&0.539&0.535&\textbf{\textcolor{blue}{0.546}}&0.542&0.548&0.543&\textbf{\textcolor{red}{0.563}}\\\tabucline[0.5pt]{-}
STMTrack&0.545&0.542&0.539&0.551&0.537&0.543&\textbf{\textcolor{blue}{0.556}}&\textbf{\textcolor{red}{0.571}}\\\tabucline[0.5pt]{-}
ARDiMP &0.543&0.551&0.554&0.572&0.547&0.564&\textbf{\textcolor{blue}{0.571}}&\textbf{\textcolor{red}{0.616}}\\\tabucline[0.5pt]{-}
TransT &0.510&0.517&0.532&0.544&0.523&\textbf{\textcolor{blue}{0.539}}&0.535&\textbf{\textcolor{red}{0.576}}\\\tabucline[0.5pt]{-}
TrSiam &0.521&0.527&0.531&0.556&0.542&0.559&\textbf{\textcolor{blue}{0.566}}&\textbf{\textcolor{red}{0.588}}\\\tabucline[0.5pt]{-}
SiamRPN &0.523&0.521&0.533&0.547&0.531&\textbf{\textcolor{blue}{0.551}}&0.543&\textbf{\textcolor{red}{0.574}}\\\tabucline[2.0pt]{-}
\end{tabu}
}
\end{center}
\label{table_newperf1}
\end{table*}

\subsubsection{UVOT Performance Comparison on Underwater Enhanced Images}
We have also compared the tracking performance of the six existing pre-trained trackers including ToMP, STMTrack, ARDiMP, TransT, TrSiam, and SiamRPN on enhanced images estimated using six existing image enhancement methods including WaterNet, FUnIEGAN, TUDA, Dive+, UColor, and Acquacolor.
We compared the performance with the raw images as well as with the proposed UWIE-TR algorithm as shown in Table \ref{table_newperf1}.
Compared to the raw images, most of the trackers have obtained better performance on the enhanced images.
It demonstrates that UWIE component improves the UVOT performance.
Among the compared trackers, ARDiMP has obtained the best success score of 61.10$\%$ using the proposed UWIE-TR algorithm.

\subsection{Evaluation using Protocol II}
For protocol II, trackers including ToMP, STMTrack, ArDiMP, TransT, TrSiam, and SiamRPN are fine-tuned on the training split of the proposed UVOT400 dataset.
These are the trackers that performed best in Protocol \textbf{I} while maintaining representation from three of the tracking categories including transformer-driven, Siamese-based, and DCFs-based.
Please note that RTS tracker is not fine-tuned since it requires a segmentation mask which is not available in the proposed dataset.
The fine-tuned trackers are evaluated using the testing split of the proposed dataset.

\subsubsection{Overall Performance}
Figs. \ref{fig5} (g)-(i) demonstrate the overall tracking performance of the selected trackers on the testing split in terms of precision, success, and normalized precision plots using OPE.
Overall, due to fine-tuning the performance of all selected trackers improved compared to Protocol \textbf{I} (Figs. \ref{fig5} (a)-(c)).
However, compared to many open-air tracking datasets such as TrackingNet, GOT-10K, and LaSOT the performance of these trackers is significantly dropped though these trackers are trained using the proposed dataset.
It is because of the additional challenges posed by the underwater environment compared to open-air sequences.

In terms of precision, ToMP has obtained the best performance of 56.70$\%$ among the selected trackers which is improved by 7.80$\%$ compared to Protocol \textbf{I}.
In terms of success, ARDiMP has obtained the best performance of 59.30$\%$ among the selected set of trackers, which is improved by 5.0$\%$ due to fine-tuning.
In terms of normalized precision, TrSiam has obtained the best performance of 69.50$\%$ which is improved by 8.20$\%$ compared to Protocol \textbf{I}.

\subsubsection{Attribute-wise Performance}
In this section, we evaluate the selected trackers on UW-specific tracking attributes in the proposed UVOT400 dataset as shown in Table \ref{table3} while the performance of all other tracking attributes is given in the supplementary material.
STMTrack tracker has obtained the best precision and normalized precision score for UWV-Low, UWV-Mid, and UWV-High attributes.
The best success score is obtained by TrSiam for UWV-LOW, STMTrack for UWV-MID, and TransT for UWV-HiGH.
Overall, STMTrack handled the target historical information and improved the tracking performance under UWV conditions.

The WCV has 16 challenging sub-attributes where different trackers are evaluated.
In colorless attribute, TransT and ToMP trackers have remained the best-performing trackers in terms of all three evaluation measures.
In Ash, ToMP has obtained the best precision and success while ARDiMP has obtained the best normalized precision and success.
In WCV-Green and LightBlue, STMTrack has obtained the best performance in terms of all measures while ToMP and ARDiMP have remained the second-best performer.
In the WCV-Blue attribute, STMTrack has remained the best performer in terms of success and ARDiMP obtained the best tracking performance in terms of precision While ToMP and STMTrack remained best in terms of normalized precision.
In WCV-Gray, ARDiMP is best in terms of precision and success while ToMP is best in terms of normalized precision.
In Cyan, TrSiam tracker obtained the best performance in all measures.
In LighGreen, ARDiMP remained the best performer while ToMP has remained the second best performer.
In DeepBlue, TrSiam has remained the best performer while TransT and SiamRPN remained the second-best performers.
In PartlyBlue, TransT has obtained the best precision and success while TrSiam obtained the best normalized precision.
Tracker ARDiMP obtained the best precision and success scores in the LightBrown attribute while TransT remained the best performer in terms of normalized precision.
In LightYellow, SiamRPN handled these sequences better than the compared trackers and remained the best performer in all three measures.
In BlueBlack, STMTrack obtained the best success and normalized precision while ARDiMP obtained the best precision.
In GrayBlue, TrSiam remained the best and TransT remained the second best-performing tracker in all three evaluation measures.

Overall, UWV-Low has remained the challenging tracking attribute for all SOTA trackers in terms of success measures.
WCV-Blue, DeepBlue, and PartlyBlue sequences have remained the challenging attribute in terms of precision and success measures.
Therefore, these four UW-specific tracking attributes posed a major challenge for compared trackers.
\textit{Compared to the UW-specific attributes, the performance of recent SOTA trackers has remained better on the attributes common with open-air trackers (please refer to Table 2 in Supplementary material).} 
\textit{The difference in performance on UW-specific and common attributes reveals the need for the development of UW-specific trackers.}

\begin{table*}[t!]
\caption{\textbf{Protocol II}: UW-specific attribute-wise performance comparison of the selected SOTA trackers in terms of Precision Rate (PR)$|$Success Rate (SR)$|$Normalized Precision Rate (NPR) on the testing split of the proposed UVOT400 dataset.
The PR is reported at a threshold of 20 pixels while AUC is shown for both SR and NPR.}
\begin{center}
\makebox[\linewidth]{
\scalebox{0.9}{
\begin{tabu}{|c|c|c|c|c|c|c|}
\tabucline[2pt]{-}
Trackers&ToMP&ARDiMP&STMTrack&TransT&TrSiam&SiamRPN\\\tabucline[2.0pt]{-}
UWV-Low (42)&0.486$|$0.443$|$0.511&0.411$|$0.402$|$0.442&\textcolor{red}{\textbf{0.512}}$|$\textcolor{blue}{\textbf{0.466}}$|$\textcolor{red}{\textbf{0.533}}&0.488$|$0.451$|$0.522&\textcolor{blue}{\textbf{0.511}}$|$\textcolor{red}{\textbf{0.488}}$|$\textcolor{blue}{\textbf{0.532}}&0.455$|$0.391$|$0.481\\\tabucline[0.5pt]{-}
UWV-Mid (41)&\textcolor{blue}{\textbf{0.533}}$|$0.500$|$\textcolor{blue}{\textbf{0.558}}&0.455$|$0.441$|$0.488&\textcolor{red}{\textbf{0.566}}$|$\textcolor{red}{\textbf{0.523}}$|$\textcolor{red}{\textbf{0.588}}&0.454$|$\textcolor{blue}{\textbf{0.501}}$|$0.488&0.512$|$0.471$|$0.551&0.499$|$0.466$|$0.521\\\tabucline[0.5pt]{-}
UWV-High (37)&0.551$|$0.523$|$0.577&0.502$|$0.482$|$0.532&\textcolor{red}{\textbf{0.603}}$|$\textcolor{blue}{\textbf{0.564}}$|$\textcolor{red}{\textbf{0.633}}&\textcolor{blue}{\textbf{0.601}}$|$\textcolor{red}{\textbf{0.588}}$|$\textcolor{blue}{\textbf{0.631}}&0.588$|$0.531$|$0.612&0.522$|$0.476$|$0.551\\\tabucline[0.5pt]{-}
WCV-Colorless (8)&\textcolor{blue}{\textbf{0.654}}$|$\textcolor{blue}{\textbf{0.611}}$|$\textcolor{red}{\textbf{0.681}}&0.622$|$0.581$|$0.644&0.631$|$0.591$|$0.622&\textcolor{red}{\textbf{0.655}}$|$\textcolor{red}{\textbf{0.623}}$|$0.661&0.632$|$0.591$|$\textcolor{blue}{\textbf{0.677}}&0.541$|$0.533$|$0.571\\\tabucline[0.5pt]{-}
WCV-Ash (2)&\textcolor{red}{\textbf{0.622}}$|$\textcolor{blue}{\textbf{0.601}}$|$\textcolor{blue}{\textbf{0.661}}&0.581$|$\textcolor{red}{\textbf{0.631}}$|$\textcolor{red}{\textbf{0.665}}&0.556$|$0.533$|$0.577&\textcolor{blue}{\textbf{0.588}}$|$0.551$|$0.601&0.561$|$0.541$|$0.591&0.491$|$0.431$|$0.523\\\tabucline[0.5pt]{-}
WCV-Green (11)&\textcolor{blue}{\textbf{0.522}}$|$0.499$|$\textcolor{blue}{\textbf{0.551}}&0.499$|$\textcolor{blue}{\textbf{0.513}}$|$0.533&\textcolor{red}{\textbf{0.531}}$|$\textcolor{red}{\textbf{0.564}}$|$\textcolor{red}{\textbf{0.601}}&0.511$|$0.491$|$0.522&0.491$|$0.476$|$0.521&0.366$|$0.351$|$0.391\\\tabucline[0.5pt]{-}
WCV-LightBlue (34)&\textcolor{blue}{\textbf{0.501}}$|$0.477$|$\textcolor{blue}{\textbf{0.533}}&0.471$|$\textcolor{blue}{\textbf{0.501}}$|$0.531&\textcolor{red}{\textbf{0.511}}$|$\textcolor{red}{\textbf{0.532}}$|$\textcolor{red}{\textbf{0.571}}&0.499$|$0.461$|$0.531&0.488$|$0.441$|$0.512&0.391$|$0.371$|$0.423\\\tabucline[0.5pt]{-}
WCV-Blue (27)&\textcolor{blue}{\textbf{0.466}}$|$\textcolor{blue}{\textbf{0.429}}$|$\textcolor{red}{\textbf{0.501}}&\textcolor{red}{\textbf{0.493}}$|$0.441$|$\textcolor{blue}{\textbf{0.488}}&0.431$|$\textcolor{red}{\textbf{0.466}}$|$\textcolor{red}{\textbf{0.501}}&0.455$|$0.422$|$0.483&0.433$|$0.402$|$\textcolor{blue}{\textbf{0.488}}&0.333$|$0.300$|$0.366\\\tabucline[0.5pt]{-}
WCV-Gray (2)&\textcolor{blue}{\textbf{0.533}}$|$\textcolor{blue}{\textbf{0.502}}$|$\textcolor{red}{\textbf{0.588}}&\textcolor{red}{\textbf{0.571}}$|$\textcolor{red}{\textbf{0.531}}$|$\textcolor{blue}{\textbf{0.561}}&0.477$|$\textcolor{blue}{\textbf{0.502}}$|$0.551&0.488$|$0.451$|$0.532&0.521$|$0.491$|$0.551&0.481$|$0.501$|$0.512\\\tabucline[0.5pt]{-}
WCV-Cyan (7)&0.588$|$0.545$|$\textcolor{blue}{\textbf{0.633}}&\textcolor{blue}{\textbf{0.601}}$|$0.551$|$0.591&0.499$|$0.523$|$0.581&0.599$|$\textcolor{blue}{\textbf{0.561}}$|$0.632&\textcolor{red}{\textbf{0.621}}$|$\textcolor{red}{\textbf{0.581}}$|$\textcolor{red}{\textbf{0.654}}&0.522$|$0.511$|$0.561\\\tabucline[0.5pt]{-}
WCV-LightGreen (4)&\textcolor{blue}{\textbf{0.544}}$|$\textcolor{blue}{\textbf{0.523}}$|$\textcolor{blue}{\textbf{0.581}}&\textcolor{red}{\textbf{0.581}}$|$\textcolor{red}{\textbf{0.541}}$|$\textcolor{red}{\textbf{0.588}}&0.455$|$0.481$|$0.522&0.499$|$0.477$|$0.521&0.531$|$0.503$|$0.561&0.422$|$0.391$|$0.455\\\tabucline[0.5pt]{-}
WCV-DeepBlue (12)&0.391$|$0.425$|$0.422&0.355$|$0.371$|$0.421&0.402$|$0.371$|$0.411&\textcolor{blue}{\textbf{0.455}}$|$0.401$|$\textcolor{blue}{\textbf{0.491}}&\textcolor{red}{\textbf{0.488}}$|$\textcolor{red}{\textbf{0.442}}$|$\textcolor{red}{\textbf{0.511}}&0.441$|$\textcolor{blue}{\textbf{0.433}}$|$0.481\\\tabucline[0.5pt]{-}
WCV-PartlyBlue (4)&\textcolor{blue}{\textbf{0.466}}$|$\textcolor{blue}{\textbf{0.441}}$|$0.499&0.431$|$0.401$|$0.451&\textcolor{blue}{\textbf{0.466}}$|$0.411$|$0.441&\textcolor{red}{\textbf{0.481}}$|$\textcolor{red}{\textbf{0.451}}$|$\textcolor{blue}{\textbf{0.522}}&0.464$|$0.410$|$\textcolor{red}{\textbf{0.551}}&0.433$|$0.402$|$0.455\\\tabucline[0.5pt]{-}
WCV-LightBrown (3)&0.501$|$\textcolor{blue}{\textbf{0.488}}$|$0.521&\textcolor{red}{\textbf{0.531}}$|$\textcolor{red}{\textbf{0.523}}$|$\textcolor{blue}{\textbf{0.541}}&\textcolor{blue}{\textbf{0.513}}$|$0.471$|$0.512&0.512$|$0.481$|$\textcolor{red}{\textbf{0.551}}&0.501$|$0.481$|$\textcolor{blue}{\textbf{0.541}}&0.455$|$0.423$|$0.471\\\tabucline[0.5pt]{-}
WCV-LightYellow (2)&0.477$|$0.441$|$0.509&\textcolor{blue}{\textbf{0.502}}$|$\textcolor{blue}{\textbf{0.461}}$|$0.496&0.491$|$0.466$|$0.511&0.477$|$0.441$|$0.522&0.491$|$\textcolor{blue}{\textbf{0.461}}$|$\textcolor{blue}{\textbf{0.533}}&\textcolor{red}{\textbf{0.561}}$|$\textcolor{red}{\textbf{0.536}}$|$\textcolor{red}{\textbf{0.611}}\\\tabucline[0.5pt]{-}
WCV-BlueBlack (1)&0.567$|$0.547$|$0.597&\textcolor{red}{\textbf{0.602}}$|$\textcolor{blue}{\textbf{0.561}}$|$\textcolor{blue}{\textbf{0.602}}&\textcolor{blue}{\textbf{0.599}}$|$\textcolor{red}{\textbf{0.577}}$|$\textcolor{red}{\textbf{0.611}}&0.501$|$0.481$|$0.521&0.488$|$0.455$|$0.511&0.464$|$0.423$|$0.523\\\tabucline[0.5pt]{-}
WCV-GrayBlue (3)&0.511$|$0.491$|$0.533&0.481$|$0.471$|$0.511&0.522$|$0.491$|$0.522&\textcolor{blue}{\textbf{0.551}}$|$\textcolor{blue}{\textbf{0.523}}$|$\textcolor{blue}{\textbf{0.581}}&\textcolor{red}{\textbf{0.588}}$|$\textcolor{red}{\textbf{0.566}}$|$\textcolor{red}{\textbf{0.622}}&0.522$|$0.481$|$0.577\\\tabucline[2.0pt]{-}
\end{tabu}
}
}
\end{center}
\label{table3}
\end{table*}

\subsection{Evaluations using Protocol III}
In this protocol, selected trackers are trained on the enhanced version of the UVOT400 dataset and tested on the unseen enhanced testing split similar to \textbf{Protocol II}.
Figs. \ref{fig5} (j)-(l) shows the results of the six SOTA trackers using \textbf{Protocol III}.
In terms of precision, ToMP has obtained the best score of 60.80$\%$ and ARDiMP has obtained the second best score of 59.80$\%$.
In terms of success rate, ARDiMP obtained the best score of 64.10$\%$ while TrSiam obtained the second-best score of 61.60$\%$.
ToMP has obtained a competitive success rate of 61.40$\%$.
Similarly, in terms of normalized precision, ToMP has obtained the best score of 73.60$\%$ while TrSiam has obtained a very competitive score of 73.40$\%$.
Compared to \textbf{Protocol II}, a ToMP tracker has obtained 4.10$\%$ better precision.
In terms of success, ARDiMP has obtained 4.80$\%$ better performance compared to \textbf{Protocol II}.
In terms of normalized precision, TrSiam demonstrated 4.10$\%$ improvement using \textbf{Protocol III}.
\textit{These significant improvements may be attributed to the UWIE-TR component proposed in Sec. \ref{sec:UIE}.
A comprehensive comparison between the six compared trackers over three protocols reveals performance improvement for each tracker which demonstrates the effectiveness of the proposed UWIE algorithm for VOT.}

\subsubsection{Attribute wise Performance}
The UW-specific attribute-wise performance using \textbf{Protocol III} is reported in Table \ref{table4} while the performance of other attributes is discussed in the supplementary material.
In the case of UWV-low, TrSiam has obtained the best performance in terms of all three measures including 55.30$\%$ precision, 53.50$\%$ success, and 57.90$\%$ normalized precision.
For UWV-Mid, ToMP has obtained the best precision score of 59.20$\%$ and STMTrack has obtained the best success and normalized precision as 56.10$\%$ and 62.20$\%$, respectively.
In the case of UWV-High, again TrSiam obtained the best precision of 64.40$\%$ while TransT obtained the best success and normalized precision of 62.10$\%$ and 67.40$\%$, respectively.

In the case of the WCV-Colorless attribute, TransT obtained the best precision and success scores of 67.80$\%$ and 64.90$\%$ while ToMP obtained the best-normalized precision of 72.20$\%$.
For the WCV-Ash, ToMP has obtained the best performance in terms of all three measures including 66.10$\%$ precision, 67.10$\%$ success, and 69.30$\%$ normalized precision.
In terms of WCV-LightBlue and WCV-Blue, STMTrack has obtained the best performance in terms of all three measures.
STMTrack also obtained success and normalized precision for the WCV-Blue attribute while ARDiMP tracker obtained the best precision score in this category.
ARDiMP further remained the dominant tracker in the presence of sequences involving WCV-Gray, WCV-Cyan, WCV-LightGreen, and WCV-LightBrown attributes.
For the WCV-DeepBlue and WCV-GrayBlue attributes, TrSiam has obtained the best scores in all three measures.
For the WCV-PartlyBlue attribute, TransT obtained the best precision and success while TrSiam obtained the best normalized precision.
For the WCV-LightYellow attribute, SiamRPN has obtained the best performance in all measures.
For the sequences involving the WCV-BlueBlack tracking challenge, STMTrack obtained the best success and normalized precision while ARDiMP obtained the best precision score.

Overall, compared to \textbf{Protocol II} (Table \ref{table3}), the performance of UWV-High attribute increased by 4.07$\%$ in terms of precision, 3.30$\%$ in terms of success, 4.10$\%$ in terms of normalized precision. 
For the WCV-Ash attribute, the performance increased by 3.90$\%$ in terms of precision, 4.00$\%$ in terms of success, and 2.80$\%$ in terms of normalized precision.
Overall, all compared trackers have obtained improved performance in \textbf{Protocol III} compared to \textbf{Protocol II}.
It demonstrates the effectiveness of our proposed UWIE-TR algorithm.


\subsubsection{UVOT Performance Comparison with Different UWIE Methods}
Using different UWIE methods, UVOT400 dataset is enhanced and six existing SOTA visual trackers are fine-tuned on the enhanced images.
Table \ref{table_newperf2} shows the performance comparisons of 48 experiments involving six existing SOTA UWIE methods, raw images, and the proposed UWIE-TR algorithm.
Compared to the raw images, all of the compared trackers have obtained better performance when fine-tuned on the enhanced images.
The performance of all trackers on the enhanced images estimated using the proposed UWIE-TR algorithm has consistently remained the best.
In most cases, TUDA has obtained the second-best performance.
This demonstrates the superiority of the proposed UWIE-TR algorithm over the existing SOTA methods for the suitability of UVOT.

\begin{table*}[t!]
\caption{\textbf{Protocol III}: Performance of fine-tuned trackers in terms of success rate. Underwater images enhanced by published enhancement methods and the proposed UWIE-TR method (right column). For all tested trackers, UWIE-TR has the largest performance gain (marked red; second best in blue).}
\begin{center}
\makebox[\linewidth]{
\begin{tabu}{|[2.0pt]c|c|c|c|c|c|c|c|c|[2.0pt]}
\tabucline[2.0pt]{-}
Trackers&RawImages&WaterNet&FUnIEGAN&TUDA&UColor&Dive+&AcquaColor&Prop. UWIE-TR\\\tabucline[0.5pt]{-}
ToMP&0.572&0.581&0.585&\textbf{\textcolor{blue}{0.596}}&0.570&0.595&0.591&\textbf{\textcolor{red}{0.614}}\\\tabucline[0.5pt]{-}
STMTrack&0.563&0.570&0.574&\textbf{\textcolor{blue}{0.584}}&0.581&0.556&0.583&\textbf{\textcolor{red}{0.604}}\\\tabucline[0.5pt]{-}
ARDiMP&0.593&0.596&0.601&\textbf{\textcolor{blue}{0.618}}&0.595&0.599&0.604&\textbf{\textcolor{red}{0.641}}\\\tabucline[0.5pt]{-}
TransT&0.559&0.564&0.571&0.580&0.542&\textbf{\textcolor{blue}{0.581}}&0.577&\textbf{\textcolor{red}{0.602}}\\\tabucline[0.5pt]{-}
TrSiam&0.571&0.574&0.579&\textbf{\textcolor{blue}{0.591}}&0.574&0.585&0.581&\textbf{\textcolor{red}{0.616}}\\\tabucline[0.5pt]{-}
SiamRPN&0.554&0.551&0.557&\textbf{\textcolor{blue}{0.571}}&0.555&0.564&0.569&\textbf{\textcolor{red}{0.585}}\\\tabucline[2.0pt]{-}
\end{tabu}
}
\end{center}
\label{table_newperf2}
\end{table*}

\subsection{Running Time}
The running time is reported using the same platform for all protocols in terms of frames per second (fps).
For \textbf{Protocols I \& II}, running time remains the same while for \textbf{Protocol III}, due to the overhead of the UWIE-TR component, the fps decreased for all compared trackers.
For the ToMP tracker, the fps decreased from 25fps for \textbf{Protocols I \& II} to 15 fps for \textbf{Protocol III}.
Similarly, for ARDiMP, STMTrack, TransT, TrSiam, and SiamRPN trackers, fps reduced from \{49, 41, 55, 38, 168\} to \{38, 35, 40, 31, 144\}, respectively.
On average, the fps were reduced by 12.50 due to the overhead of the image enhancement step. 
All trackers still remain real-time trackers,  excluding ToMP.

\section{Discussion \& Conclusion}
\label{sec:conclusion}
In this work, we proposed a new large-scale UVOT dataset consisting of 400 video sequences and 275,000 annotated frames.
Underwater tracking presents several unique challenges compared to open-air VOT such as UW image quality degradation, poor lighting conditions, varying UW colors, object appearance variations, and UW sensor limitations.
We show that the performance of many existing SOTA trackers drops significantly on our proposed UVOT dataset mainly due to the underwater inherited attributes.
We have also presented a generic underwater image enhancement algorithm as a pre-processing step for existing SOTA visual trackers.
We further demonstrate that UVOT performance improves using the enhanced images estimated by the proposed UWIE-TR algorithm.

Experiments conducted using three different protocols demonstrated the challenging nature of our proposed dataset.
In \textbf{protocol I}, we evaluated open-air trackers on the proposed UVOT 400 dataset with no re-training and observed performance degradation of up to 26.90$\%$.
In \textbf{protocol II}, when open-air trackers were fined-tuned on the training split of the UVOT dataset, the performance of all trackers improved but still remained significantly lesser than the open-air performance.
In \textbf{protocol III}, existing SOTA trackers are evaluated on the enhanced images of the UVOT dataset estimated by our proposed UWIE-TR algorithm.

The main challenge in developing the UWIE-TR algorithm is the lack of ground truth paired images.
We handled this challenge by using multiple UW image enhancement methods and then manually selected the best-enhanced images as the ground truth pair to train our proposed algorithm.
Our UWIE algorithm consists of a features extraction head, UWTE followed by an enhanced image decoder.
Latent space information fusion is employed to obtain performance improvement.
The proposed UWIE algorithm is compared with existing SOTA UW image enhancement methods on publicly available datasets.
Our results demonstrate significant performance improvement using no-reference image quality measures as well as reference-based UW measures.

Compared to the results on the raw UVOT400 sequences, the performance of SOTA trackers improved up to 5.00 $\%$. 
Our proposed UWIE-TR algorithm resulted in performance improvement of most of the existing SOTA trackers.
For instance, for the ToMP tracker, the performance in terms of success rate on the TrackingNet dataset is 81.50$\%$ while the performance on the raw UVOT dataset is 52.60$\%$.
The performance of the same tracker on enhanced sequences is 61.40$\%$.
Similarly, for the STMTrack tracker, the performance on the TrackingNet dataset is 80.30$\%$ while the performance on raw UVOT sequences is 54.50$\%$.
The image enhancement of UVOT sequences resulted in 60.40$\%$ performance.
Thus, the performance gap between the open-air and UW sequences is reduced up to 6.0$\%$ while there is still a gap of almost 20.0$\%$.
In the future, underwater-specific trackers are needed which should be able to enhance the underwater raw images and perform robust tracking against distractors in an end-to-end manner.

\begin{table*}[t!]
\caption{\textbf{Protocol III}: UW-specific attribute-wise performance comparison of the selected SOTA trackers in terms of Precision Rate (PR)$|$Success Rate (SR)$|$Normalized Precision Rate (NPR) on the testing split of the proposed UVOT400 dataset.
The PR is reported at a threshold of 20 pixels while AUC is shown for both SR and NPR.}
\begin{center}
\makebox[\linewidth]{
\scalebox{0.9}{
\begin{tabu}{|c|c|c|c|c|c|c|}
\tabucline[2pt]{-}
Trackers&ToMP&ARDiMP&STMTrack&TransT&TrSiam&SiamRPN\\\tabucline[2.0pt]{-}
UWV-Low (42)&0.521$|$0.471$|$0.566&0.458$|$0.438$|$0.495&\textcolor{blue}{\textbf{0.544}}$|$\textcolor{blue}{\textbf{0.505}}$|$\textcolor{blue}{\textbf{0.574}}&0.525$|$0.485$|$0.551&\textcolor{red}{\textbf{0.553}}$|$\textcolor{red}{\textbf{0.535}}$|$\textcolor{red}{\textbf{0.579}}&0.491$|$0.422$|$0.542\\\tabucline[0.5pt]{-}
UWV-Mid (41)&\textcolor{red}{\textbf{0.592}}$|$0.542$|$\textcolor{blue}{\textbf{0.592}}&0.481$|$0.472$|$0.532&\textcolor{blue}{\textbf{0.581}}$|$\textcolor{red}{\textbf{0.561}}$|$\textcolor{red}{\textbf{0.622}}&0.481$|$\textcolor{blue}{\textbf{0.532}}$|$0.521&0.536$|$0.515$|$0.572&0.529$|$0.501$|$0.555\\\tabucline[0.5pt]{-}
UWV-High (37)&0.592$|$0.572$|$0.611&0.535$|$0.527$|$0.572&0.632$|$\textcolor{blue}{\textbf{0.603}}$|$\textcolor{blue}{\textbf{0.666}}&\textcolor{blue}{\textbf{0.633}}$|$\textcolor{red}{\textbf{0.621}}$|$\textcolor{red}{\textbf{0.674}}&\textcolor{red}{\textbf{0.644}}$|$0.565$|$0.638&0.558$|$0.522$|$0.589\\\tabucline[0.5pt]{-}
WCV-Colorless (8)&\textcolor{blue}{\textbf{0.671}}$|$\textcolor{blue}{\textbf{0.632}}$|$\textcolor{red}{\textbf{0.722}}&0.651$|$0.633$|$0.675&0.653$|$0.625$|$0.644&\textcolor{red}{\textbf{0.678}}$|$\textcolor{red}{\textbf{0.649}}$|$0.693&0.655$|$0.623$|$\textcolor{blue}{\textbf{0.702}}&0.573$|$0.566$|$0.605\\\tabucline[0.5pt]{-}
WCV-Ash (2)&\textcolor{red}{\textbf{0.661}}$|$\textcolor{red}{\textbf{0.671}}$|$\textcolor{red}{\textbf{0.693}}&0.616$|$\textcolor{blue}{\textbf{0.668}}$|$\textcolor{blue}{\textbf{0.691}}&0.583$|$0.562$|$0.601&\textcolor{blue}{\textbf{0.602}}$|$0.598$|$0.635&0.595$|$0.567$|$0.645&0.537$|$0.477$|$0.554\\\tabucline[0.5pt]{-}
WCV-Green (11)&\textcolor{blue}{\textbf{0.557}}$|$0.533$|$\textcolor{blue}{\textbf{0.584}}&0.524$|$\textcolor{blue}{\textbf{0.557}}$|$0.574&\textcolor{red}{\textbf{0.566}}$|$\textcolor{red}{\textbf{0.592}}$|$\textcolor{red}{\textbf{0.634}}&\textcolor{blue}{\textbf{0.557}}$|$0.533$|$0.567&0.532$|$0.518$|$0.563&0.401$|$0.398$|$0.435\\\tabucline[0.5pt]{-}
WCV-LightBlue (34)&0.532$|$0.512$|$0.571&0.525$|$\textcolor{blue}{\textbf{0.544}}$|$0.566&\textcolor{red}{\textbf{0.536}}$|$\textcolor{red}{\textbf{0.566}}$|$\textcolor{red}{\textbf{0.606}}&\textcolor{blue}{\textbf{0.533}}$|$0.495$|$\textcolor{blue}{\textbf{0.583}}&0.522$|$0.484$|$0.555&0.422$|$0.402$|$0.460\\\tabucline[0.5pt]{-}
WCV-Blue (27)&\textcolor{blue}{\textbf{0.505}}$|$0.445$|$\textcolor{blue}{\textbf{0.530}}&\textcolor{red}{\textbf{0.526}}$|$\textcolor{blue}{\textbf{0.475}}$|$0.524&0.466$|$\textcolor{red}{\textbf{0.492}}$|$\textcolor{red}{\textbf{0.533}}&0.491$|$0.465$|$0.525&0.481$|$0.446$|$0.522&0.361$|$0.344$|$0.391\\\tabucline[0.5pt]{-}
WCV-Gray (2)&0.553$|$0.535$|$\textcolor{red}{\textbf{0.625}}&\textcolor{red}{\textbf{0.611}}$|$\textcolor{red}{\textbf{0.566}}$|$0.592&0.515$|$\textcolor{blue}{\textbf{0.545}}$|$\textcolor{blue}{\textbf{0.599}}&0.528$|$0.485$|$0.565&\textcolor{blue}{\textbf{0.558}}$|$0.533$|$0.588&0.522$|$0.533$|$0.545\\\tabucline[0.5pt]{-}
WCV-Cyan (7)&0.611$|$0.571$|$0.665&\textcolor{red}{\textbf{0.644}}$|$0.592$|$0.633&0.522$|$0.561$|$0.622&0.633$|$\textcolor{blue}{\textbf{0.595}}$|$\textcolor{blue}{\textbf{0.666}}&\textcolor{blue}{\textbf{0.642}}$|$\textcolor{red}{\textbf{0.623}}$|$\textcolor{red}{\textbf{0.681}}&0.568$|$0.553$|$0.599\\\tabucline[0.5pt]{-}
WCV-LightGreen (4)&\textcolor{blue}{\textbf{0.571}}$|$\textcolor{blue}{\textbf{0.556}}$|$\textcolor{red}{\textbf{0.631}}&\textcolor{red}{\textbf{0.622}}$|$\textcolor{red}{\textbf{0.586}}$|$\textcolor{blue}{\textbf{0.626}}&0.481$|$0.522$|$0.557&0.522$|$0.517$|$0.556&0.564$|$0.535$|$0.594&0.467$|$0.425$|$0.481\\\tabucline[0.5pt]{-}
WCV-DeepBlue (12)&0.422$|$\textcolor{blue}{\textbf{0.465}}$|$0.447&0.381$|$0.414$|$0.457&0.447$|$0.424$|$0.458&\textcolor{blue}{\textbf{0.491}}$|$0.444$|$\textcolor{blue}{\textbf{0.525}}&\textcolor{red}{\textbf{0.533}}$|$\textcolor{red}{\textbf{0.483}}$|$\textcolor{red}{\textbf{0.557}}&0.467$|$0.462$|$0.521\\\tabucline[0.5pt]{-}
WCV-PartlyBlue (4)&\textcolor{blue}{\textbf{0.493}}$|$\textcolor{blue}{\textbf{0.477}}$|$0.531&0.466$|$0.434$|$0.487&0.489$|$0.445$|$0.493&\textcolor{red}{\textbf{0.522}}$|$\textcolor{red}{\textbf{0.496}}$|$\textcolor{blue}{\textbf{0.567}}&0.501$|$0.448$|$\textcolor{red}{\textbf{0.581}}&0.476$|$0.437$|$0.491\\\tabucline[0.5pt]{-}
WCV-LightBrown (3)&\textcolor{blue}{\textbf{0.548}}$|$0.519$|$0.557&\textcolor{red}{\textbf{0.584}}$|$\textcolor{red}{\textbf{0.561}}$|$\textcolor{red}{\textbf{0.583}}&0.549$|$0.517$|$0.545&0.537$|$\textcolor{blue}{\textbf{0.522}}$|$\textcolor{blue}{\textbf{0.580}}&0.543$|$0.513$|$0.567&0.471$|$0.458$|$0.516\\\tabucline[0.5pt]{-}
WCV-LightYellow (2)&0.515$|$\textcolor{blue}{\textbf{0.532}}$|$0.532&\textcolor{blue}{\textbf{0.535}}$|$0.501$|$0.531&0.525$|$0.507$|$0.545&0.504$|$0.486$|$\textcolor{blue}{\textbf{0.566}}&0.523$|$0.494$|$\textcolor{blue}{\textbf{0.562}}&\textcolor{red}{\textbf{0.594}}$|$\textcolor{red}{\textbf{0.564}}$|$\textcolor{red}{\textbf{0.632}}\\\tabucline[0.5pt]{-}
WCV-BlueBlack (1)&0.567$|$0.547$|$\textcolor{blue}{\textbf{0.605}}&\textcolor{red}{\textbf{0.602}}$|$\textcolor{blue}{\textbf{0.561}}$|$0.602&\textcolor{blue}{\textbf{0.599}}$|$\textcolor{red}{\textbf{0.577}}$|$\textcolor{red}{\textbf{0.611}}&0.501$|$0.481$|$0.521&0.488$|$0.455$|$0.511&0.464$|$0.423$|$0.523\\\tabucline[0.5pt]{-}
WCV-GrayBlue (3)&0.555$|$0.532$|$0.561&0.525$|$0.514$|$0.544&0.551$|$0.533$|$0.557&\textcolor{blue}{\textbf{0.582}}$|$\textcolor{blue}{\textbf{0.561}}$|$\textcolor{blue}{\textbf{0.606}}&\textcolor{red}{\textbf{0.614}}$|$\textcolor{red}{\textbf{0.591}}$|$\textcolor{red}{\textbf{0.661}}&0.557$|$0.515$|$0.608\\\tabucline[2.0pt]{-}
\end{tabu}
}
}
\end{center}
\label{table4}
\end{table*}

\section*{Acknowledgments} 
This publication acknowledges the support provided by the Khalifa University of Science and Technology under Faculty \textcolor{black}{Start-Up} grants FSU-2022-003 Award No. 8474000401.

\ifCLASSOPTIONcaptionsoff
  \newpage
\fi

\bibliographystyle{IEEEtranS}
\bibliography{bare_jrnl_compsoc}

\end{document}